# A Survey on Deep Domain Adaptation and Tiny Object Detection Challenges, Techniques and Datasets

MUHAMMAD MUZAMMUL and XI LI*, College of Computer Science, Zhejiang University, China

Deep learning (DL) and computer vision (CV) offered a trending role in object detection(OD), object tracking, pedestrian detection, and autonomous vehicles from the last few decades. Several old and recent approaches were proposed to solve these computer vision and deep learning-based problems with detection, tracking techniques, algorithms, and data sources. In recent decades, this field gained importance due to the rising interest in brain-inspired human recognition and detection technologies. Using online/offline data about images and videos, researchers are intensively modeling sentiments and computational analysis. Computer vision-based artificial neural networks (ANN) and convolutional neural network (CNN) based approaches providing robust solutions. This survey paper specially analyzed computer vision-based object detection challenges and solutions by different techniques. We mainly highlighted object detection by three different trending strategies, i.e., 1) domain adaptive deep learning-based approaches (discrepancy-based, Adversarial-based, Reconstruction-based, Hybrid). We examined general as well as tiny object detection-related challenges and offered solutions by historical and comparative analysis. In part 2) we mainly focused on tiny object detection techniques (multi-scale feature learning, Data augmentation, Training strategy (TS), Context-based detection, GAN-based detection). In part 3), To obtain knowledge-able findings, we discussed different object detection methods, i.e., convolutions and convolutional neural networks (CNN), pooling operations with trending types. Furthermore, we explained results with the help of some object detection algorithms, i.e., R-CNN, Fast R-CNN, Faster R-CNN, YOLO, and SSD, which are generally considered the base bone of CV, CNN, and OD. We performed comparative analysis on different datasets such as MS-COCO, PASCAL VOC07,12, and ImageNet to analyze results and present findings. At the end, we showed future directions with existing challenges of the field. In the future, OD methods and models can be analyzed for real-time object detection, tracking strategies.

CCS Concepts: • **Deep learning→Object detection**; *Challenges, Deep domain adaptation*; Tiny objects; • **Techniques**→Methods.

Additional Key Words and Phrases: datasets, models, methods, trends

## 1 INTRODUCTION

The computer vision (CV) field performed a vital role in detection and tracking. In the last few decades, trending research in this field attracted many researchers for the applications of robotics, driverless car, disabled body parts, speech recognition devices, security-based detection systems, medical tumor detection, and indexing in contents. Artificial intelligence (AI) supports solutions with the help of CV, CNN in basic/advanced detection applications. In this survey paper, we focused on trending OD tricks and methodologies with computer vision-based approaches. Several object detection methods (ODM) related to Convolution (one-to-one convolution, 3D Convolution, Coordinate Convolution & other) and CNN (Residual Network, AlexNet, VGG, DenseNet & others) and pooling based operations with several types discussed at the same place.

Different strategical analyses can classify object detection and tracking, but in this paper, our primary focus is tiny OD and general concepts of OD with the help of CV-based trending techniques. Several CV-based applications gained researchers' attention based on audio,video & image-based data for pedestrians and object detection techniques, human interactions with machines, combinational analysis of ANN, CNN, speech and text processing concerning online and offline data. Variety of data present to detect and track dimensional interfaces by 2D and 3D indoor and outdoor systems. We also considered the recent trending field of convolutional neural networks that worked as a building block for object detection and tracking with video or image-based data processing. Deep domain adaptation-based OD strategies and its types (discrepancy-based, Adversarial-based, Reconstruction-based, Hybrid) also offering a significant role in object detection that is also part of this survey. Generally, when OD is considered, tiny objects with different types of problems (discussed in section 1.6) also gained the researcher's attention. We discussed various techniques in section 2 with the help of literature that support knowledgeable solutions for tiny OD and general OD. Due to tiny OD challenges, we focused more on it concerning existing strategies from literature. We mainly focused on tiny object detection techniques (multi-scale feature learning, Data augmentation, Training strategy (TS), Context-based detection, GAN-based detection) We presented the results of different papers by applying computer vision-based techniques/Models/Methods on existing datasets, i.e., MS-COCO, PASCAL VOC07,12, and ImageNet and these are considered as highly reputed datasets for object detection purposes.

Convolution methods and deep convolutional neural networks, as well as GPU's high computing power, can be regarded as the most important contributors to these advancements [1-3]. Deep learning-based computational models are currently widely used for both generic and domain-specified object detection. These computational models serve as the backbone architecture in the majority of object detectors. They are extensively used to perform various tasks such as feature extraction from an input image, segmentation, classification, and object localization. Image classification [4,5], semantic segmentation [6], object detection, and instance segmentation [7] are all examples of recognition problems in the field of computer vision.

*Corresponding Author: xilizju@zju.edu.cn

Author's address: Muhammad Muzammul, muzamal@zju.edu.cn; XI LI, xilizju@zju.edu.cn, College of Computer Science, Zhejiang University, China, P.O. Box W-99, Hangzhou, Zhejiang, China, 310027.

Many benchmarks, i.e., Caltech [8], KITTI [9], ImageNet [10], PASCAL VOC [11], MS COCO [12], and Open Images V5 [13] performing a significant role in the object detection sector so far. The ECCV VisDrone 2018 conference organizers have published a novel drone platform-based dataset [14] with many photos and videos. A good object detection algorithm should be able to understand both semantic and spatial details about the image. A good object detection algorithm should be able to understand both semantic and spatial details about the image. Object detection has been widely supporting in computer vision applications such as face recognition [15,16,17], pedestrian detection [18,19,20], logo detection [21,22], and video analysis [23,24] as an essential part of image comprehension.

However, due to vast differences in lighting conditions and occlusions and various perspectives and poses, the process of object detection and localization has become repetitive and difficult to achieve with absolute precision. As a result of these obstacles, this sector has received much more attention in recent years [6,25,26,27].

Object detection strategies based on deep learning [5,27] have made considerable progress in recent years, outperforming conventional detectors, thanks to the advent of deep convolutional neural networks [33] and their widespread applications in image classification [5,38]. Deep CNN-based object detectors create hierarchical feature representations trained in a self-automated manner and attempt to display more discriminative speech power than conventional object detectors that use manually designed feature descriptors.

The following are the benefits of CNN-based approaches over conventional object detectors:

- Compared to shallow conventional models, a deep CNN-based model offers exponentially more expressive capacity.
- CNN-based object detectors enable you to combine multiple detection-related tasks into one.
- Hierarchical function vectors/representations can be extracted automatically from the underlying data in CNN-based object detectors and disentangled using multi-level nonlinear mappings[28-30].

All of these benefits made it clear that deep learning-based object detection techniques with expressive feature representation capability could be designed and developed in an end-to-end manner. These characteristics have enabled CNNs to be applied to a variety of research domains, including image classification [34], face recognition [35], video analysis [36,37], and pedestrian detection [38,39,40].

Tiny object detection is an essential fundamental computer technology that identifies tiny objects of a specific class in digital images and videos. It is related to image understanding and computer vision. Tiny object detection is a key and complex problem in computer vision, as it is the foundation for many other computer vision tasks such as object tracking [135], instance segmentation [136, 137], image captioning [138], action recognition [139], scene comprehension [140], and so on. In recent years, the remarkable success of deep learning techniques has infused new life into tiny object detection, propelling it to the forefront of science. Tiny object detection is widely used in academia and real-world applications, including robot vision, autonomous driving, intelligent transportation, drone scene analysis, and military reconnaissance and surveillance [31-33]. Furthermore, since most previous efforts have been tuned for the more significant object detection problem, tiny object detection experience and expertise are extremely limited. This research elaborated on the general object detection concept, domain adaptive object detection, and tiny object detection.

### 1.1. Structure of Survey

This survey is structured as follows: Section 1 contains a complete introduction to computer vision-based object detection, applications, and problems. Section 2 elaborate techniques to solve OD-related issues, object detection trending approaches, i.e., 2.1 elaborate general challenges for tiny OD and solution with literature references, 2.2 we offered more techniques using deep domain adaptation deep learning-based OD and some relationship with tiny OD. In section 2.3, we explained tiny OD and related techniques. In section 2.4, general state-of-art and 2.5 explained the relationship b/w DDA-OD and tiny-OD. Section 3 explains some architectures/Models/algorithms with a special focus on convolution, convolution neural networks, and pooling operations and types for more OD understanding. In Section 4, comparative analysis of different models, methods, and strategies helpful for OD and presented results in tabular form. In section 5, we offered a conclusion with several future research directions.

### 1.2. Motivations for research

Great strides have been made in general object detection with the advancement of deep learning and the continuous improvement of computing resources. When the first CNN-based object detector, R-CNN, was proposed, it sparked a series of significant contributions that have significantly offered excellent help for advancements in general object detection and tiny object detection. To help beginners get started in this field, we present some object detection architectures, different trending strategies with the help of literature trends, and comparative analysis in tabular form.

Table 1: Summery of most popular object detection (OD) surveys Table 2.

| No. | Tittle/Area focused | Year | Explanation | Ref # |
|---|---|---|---|---|
| 1 | Deep learning (DL) | 2013<br>2015<br>2017<br>2018 | These surveys focused on the particular field of artificial intelligence that works according to the brain rule of data processing and helps in object detection, speech recognition, language translation, and decision support systems. With deep learning AI work without human support. It gets decisions from unlabeled and unstructured data. [ tutorials at ICCV and CVPR] | [56]<br>[57]<br>[58]<br>[59] |
| 2 | Traffic sign detection | 2016 | Traffic sign detection also important for traffic rules control | [62] |
| 3 | Tiny object detection | 2021 | A survey and performance evaluation of deep learning methods for small object detection | [81] |
| 4 | Deep learning | 2015 | An introduction to deep learning and applications | [82] |
| 5 | A survey on deep learning in medical image analysis | 2017 | A survey on deep learning-based image classification, object detection, segmentation, and registration in medical image analysis | [83] |
| 6 | Recent advances in convolutional neural networks | 2015 | An introduction to deep learning and applications | [82] |
| 7 | Deep learning | 2017 | A large survey on advancements in convolutional neural networks(CNN) and their practical applications in computer vision, speech recognition, and natural language processing | [84] |
| 8 | Object detection in 20 years: A survey | 2019 | A comprehensive review on the technical evolution of different object detection techniques in the last two decades. Given an overview of recent and old object detection approaches. | [85] |
| 9 | A survey of deep learning-based object detection | 2019 | A survey with focus on analyzing deep learning-based object detection task | [86] |
| 10 | Deep learning | 2019 | A systematic review to summarize representative models and their different characteristics in several generic object detection application domains | [87] |
| 11 | Recent advances in deep learning for object detection | 2020 | Presents a comprehensive understanding of deep learning-based object detection algorithms | [88] |

### 1.3. Relevance to other surveys and significance

Object detection is a vast field, and several surveys are published due to the trending attention of computer vision-based humanoid devices. In Table 1, we presented some old and recent surveys related to computer vision and object detection. The researches that we presented here given us the idea to write our survey paper. (Andreopoulos et al. 2013)50 years of object recognition: directions forward published in 2013 CVIU focus on "A review of the evolution of object recognition systems over 5 decades". This comprehensive survey has a great role in object detection, and we also get the idea after getting the historical directions field [75]. Zhang et al. (2013) Object class detection: a survey published in 2013 ACM CS focused on "Survey of generic object detection methods before 2011" [76]. Bengio et al. 2013 in PAMI contributed to unsupervised feature learning and deep learning, probabilistic models, autoencoders, manifold learning, and deep networks. Salient object detection: a survey [77].(Borji et al. 2014), arXiv, A survey of salient object detection techniques. A survey on face detection in the wild: past, present, and future [78].(Zafeiriou et al., 2015), CVIU, A survey of face detection techniques in the wild since 2000.Text detection and recognition in imagery: a survey [79].(Ye et al. 2015), PAMI, A thorough review of text detection and recognition in colored images. Feature representation for statistical learning-based object detection: a review [80]. (YangLiu et al., 2021) published a survey paper with a title, a survey, and a performance evaluation of deep learning methods for small object detection [81] that focused on challenges in tiny object detection and provided some techniques to solve these problems. (LeCun et al. 2015), Nature, An introduction to deep learning and applications. A survey on deep learning in medical image analysis [82]. (Zafeiriou et al., 2015) A survey on face detection in the wild: past, present and future published in 2015 CVIU with focus on" A survey of face detection in the wild since 2000" [55]. (Ye et al., 2015) Text detection and recognition in imagery: a survey published in 2015 IEEE PAMI special focus on "A survey of text detection and recognition in color imagery" [57]. (Zhengxia Zou et al.2019) Object Detection in 20 Years: A Survey published in computer vision and Pattern Recognition with special focus on "historical background of object detection methods as well as to object detection models that proposed in different object detection branches" [60], This was a comprehensive survey on OD that given us beneficial knowledge for field study. (Litjens et al. 2017), MIA, A survey on deep learning-based image classification, object detection, segmentation and registration in medical image analysis. Recent advances in convolutional neural networks. (Gu et al.2015), Nature,2015, An introduction to deep learning and applications. Deep learning [84]. (Zou et al. 2019), IEEE,2019, A comprehensive review in the light of technical evolution of different object detection techniques in last two decades. A survey of deep learning-based object detection [85]. (Jiao et al.2019), IEEE,2019, A broad survey focuses on describing and analyzing deep learning-based object detection tasks. Deep learning [86]. (LeCun et al.2019),arXiv, 2019, A systematic review summarises representative models and their different characteristics in several generic object detection application domains. Recent advances in deep learning for object detection [87]. (Wu et al.2020), Neurocomputing,2020, Presents a comprehensive understanding of deep learning-based object detection algorithms [88]. In general, we performed analytical review about different contributions with relevance deep learning trending fields such as pedestrian detection (PD)[41-43][53], face detection(FD) [44], text Detection(TD) [47], vehicle detection(VD) [46][42], generic object detection(GOD)[53],DCNNS, Image, Video, Speech and Audio processing [63-65], deep learning and domain adaptive object detection (DL & DAOD) [56-59].traffic sign detection [62]. From [62-83], different surveys related to general object detection, tiny object detection, and deep domain adaptive object detection we studied to develop our survey paper.

### 1.4. Contribution of this survey

The primary goal of this survey is to identify and evaluate the object detection problem using deep learning. Unlike previous surveys, this paper discussed state-of-the-art object detection techniques and various learning methods. This research-based survey includes an in-depth review and comprehensive discussion of various topics, some of which are novel to our understanding. We discussed some trending object detection techniques like domain adaptive deep learning, tiny object detection state-of-the-art techniques and analyzed results with the help of different object detection methods, models, and benchmarks.

- We have mentioned some object detection learning strategies in this paper and have not included detailed basic information; however, we have addressed recent, old object detection trends so that readers can easily get depth knowledge at the same place.
- Unlike previous surveys in this area, this paper offers a thorough and systematic examination of deep learning-based object detection techniques, as well as a summary of the most crucial research trends and current object detection algorithms. Most of the results we displayed in tabular form with comparative analysis and literature reference. Readers can get the idea of methodology quickly with tiny insight.
- This survey paper is unique because different computer vision and essential object detection strategies/directions are reviewed simultaneously. Each section offering a complete idea about object detection and results proved with comparative analysis.
- This survey paper consists of well-structured computer vision & object detection; it consists of 567 references and 16 tables. All tables give comparative analysis or literature ideas about trending object detection methodologies or results. Table 2 included(>125 literature references) offering a great idea about computer vision-based OD applications with trending problems and future research direction.
- We specially analyzed object detection applications and surveys in different fields in Table 1, Table 2. Section 2 discussed OD trending approaches such as deep domain adaptive object detection (DDAOD) with varying OD terminologies. This subject is addressed very short in literature, and it's not so common topic. The result of comparative analysis on different OD methods and datasets about DDAOD shown in Table 3(3.1,3.2,3.3,3.4,3.5).
- Next, we highlighted the second domain with tiny object detection approaches as presented different methods with the historical and taxonomical point of view in Table 4,5.
- Section 3 explains object detection methods and types of these methods, focusing on Convolution, convolution neural networks, and pooling operations with a combined picture of many techniques at the same place.
- Table 7,8,9 in section 4 contributed with detailed analysis of object detection backbone models, i.e., CNN, R-CNN, Fast R-CNN, YOLO, SSD and comparative results as pros and cons presented. After performing experiments on datasets with OD approaches/ Methods/Models, the final results are presented in Table 10-16.
- In section 5, we offered a comprehensive conclusion and future direction.

As a result, this survey paper aims to provide a comprehensive study of deep learning-based object detection/challenges/applications and trending techniques to solve tiny OD and DDA-OD issues. In the end, we have given conceptual ideas for future directions to readers.

Table 2. Computer vision/Object detection methods and applications (Overview)

| Previous work | Research Problems & directions | Background literature | Traditional common solution methods | Deep learning-based solution methods | Short explanation and future directions |
|---|---|---|---|---|---|
| | | 1.5.1. Pedestrian Detection (PD) | | | |
| HOG detector [373], ICF detector [374], Feature representation [373,374], Design classifier [375], Detection acceleration [376], | 1)Small pedestrian, 2)Hard negatives, 3)Dense and occluded pedestrian, 4)Real-time detection | Long research history [383,384]. Two special ways 1) Traditional PD, 2) Deep Learning-based PD Need to follow surveys [377, 378-382]. | Early time PD [383-385,] "Detection by components" [384, 386,387], gradient-based representation [373, 387-390] DPM [390-392] (ICF) [374]. new benchmark of PD [377]. Shape symmetry [310] and stereo information [173, 311]. | Improve small PD: Deep learning [395]. Convolutional features [396]. Feature fusion [396], Handcrafted features [397,398], Multiple resolutions [399]. Negative detection [396] [400][401] Dense and occluded PD [402-405] | Although deep learning object detectors such as Fast/Faster R-CNN performed a great role as general object detectors [406] for pedestrian detection [396]. Still, this field is getting many new methods in deep learning CNN due to different problems in OD. |
| | | 1.5.2. Face Detection (FD) | | | |
| Oldest computer vision applications [407,408] VJ detector [409] | 1)Intra-class variation, 2) Occlusion faces, 3)Multi-scale detection, 4)Real-time detection | Early 1990s: [410,411][413] Before 2001: Relationship's b/w facial elements [412,413]. Face distribution in subspace [410,411] | In 2000-2015: Boosted decision trees [409,414, 109], Speed up detection [112, 113, 415] | Speed up face detection: [503,504]. Improve multi-pose and occluded face detection: [419-422] | General object detectors such as Faster R-CNN and SSD are working these days as detection base bone. Several new methods are expected in the future. |
| | | 1.5.3. Text Detection (TD) | | | |
| Street sign and currency reading [423,424] Digital map Build [425,426] | 1)Different fonts and languages, 2)Text rotation and perspective distortion, 3)Densely arranged text localization, 4) Broken and blurred characters | Text detection consists of two steps: 1) Text localization, 2) Text recognition. Literature offered "step-wise detection" and "integrated detection" methods. Need to follow survey: [427,428] | Step-wise detection methods [429,430] By contrast integrated methods [431-434] candidate windows [428] Maximally Stable Extremal Regions (MSER) segmentation [430] Morphological filtering [435] Texts and the structures of strokes [429,430, 436]. | General object detection [437, 438-447] Segmentation problem [448, 449, 451,452]. For text rotation and perspective changes: [441-442,345–447]. Densely arranged text detection: [448]. Individual line of text [449, 451,452]. Broken and blurred text detection [453,454] [445] [453, 438]. | Computer vision, i.e., supervised and unsupervised object detection techniques, also getting trends for future research. Many scientists contributed and still need a lot of work on it for special autonomous text reader applications. |
| | | 1.5.4. Traffic Sign and Traffic Light Detection (TSTLD) | | | |
| Fixed patterns recognition like traffic signs detection remained a less focused area at an early age, and more focus is gained by computer vision & deep learning-based applications | 1)Illumination changes, 2)Motion blur, 3)Bad weather, 4)Real-time detection, | Divided into two groups: 1)Traditional detection methods, 2)Deep learning-based detection methods Need to follow survey: [455] | Traffic sign/light detection 20 years ago [456, 457]. Color thresholding [458–463], Visual saliency detection [463], Morphological filtering [505], Edge/contour analysis [464,465]. GPS and digital maps in traffic light detection [466,467]. | In the deep learning era, some well-known detectors such as Faster RCNN and SSD were applied in traffic sign/light detection tasks [468, 468, 470,471]. Adversarial training has been used to improve detection under complex traffic environments [472, 470]. | General object detection and tracking get significant attention from computer vision and the deep learning community. So moving and static object detection and tracking is the essential future of an autonomous environment. |
| | | 1.5.5. Remote Sensing Target Detection (RSTD) | | | |
| RSTD (e.g., the detection of airplane, ship, oil-pot, etc.). RSTD applications in (military investigation, disaster rescue, and urban traffic management) | 1)Detection using "big data." 2)Occluded targets, 3)Domain adaptation | following surveys for more details on this topic [473,474] | RSTD is a two-stage detection paradigm: 1) Candidate extraction, 2) Target verification. In methods include gray Value filtering [475,476], Visual saliency [477-480], Wavelet transform [481], Anomaly detection [482]. In target verification stage, some frequently used features include HOG [482,483], LBP [476], SIFT [478, 480, 484]. Sliding window detection paradigm [483-486] | After success of R-CNN deep CNN based methods performed great role in RSTD [487,488, 489]. SSD have attracted increasing attention in remote sensing community [490-495]. Deep CNN features for remote sensing images [496-498]. People. ROI Pooling layer for better rotation invariance [499,500]. strategy used to improve small target detection [501,502]. | In recent years, as the resolution of remote sensing images has increased, After the great success of RCNN in 2014, deep CNN has been soon applied to remote sensing target detection. Still, this field attracting outstanding research contributions due to its success in considerable data resources. |

## 1.5. Computer vision-based object detection applications (Overview of field)

This introduction will highlight a basic short description of the field and five major computer vision applications in which all research communities offer day-by-day new research contributions. Instead of providing details, we would like to present data in Table 2; we hope that all detection ideas will be cleared by visualizing tabular information. Before further explanation, we would like to mention significant CV applications; these applications have different problems and have other solutions.

### *1.5.1. Pedestrian detection (PD)*

Pedestrian detection has received much exposure and research contributions in recent years due to critical object detection applications, including autonomous driving vehicles and humanoid devices, video-based monitoring, and criminal investigation. Pedestrian detection methods from the past, such as HOG detector [373], ICF detector [374], feature representation [373,374], design classifier [375], detection acceleration [376]. This field has different problematic areas (small pedestrian, hard negatives, dense and occluded pedestrian, real-time detection). Different researchers showed great contributions in literature and many new concepts causing new contributions recently and expected in future research. We presented an overview in Table 2.

### *1.5.2. Face detection (FD)*

Face recognition-based applications are typically considered one of the earliest uses of computer vision [407,408]. Face detection from the beginning, such as the VJ detector [409], has dramatically aided object detection, many of its groundbreaking ideas still playing essential roles in today's object detection. Face recognition is now used in almost every aspect of life, including digital camera "facial expression" detection, e-commerce "face swiping," mobile app facial makeup, and so on. This field has different problematic directions (intra-class variation, occlusion faces, multi-scale detection, real-time detection). For literature reference and additional existing contribution overview, please see Table 2.

### 1.5.3. *Text detection (TD)*

From several decades, the text remains a primary source of communication between humans. The main objective of text detection is to decide whether there is text in a given image or not. In short, we can say recognizing and locating text inside an image according to need and requirements considered as TD in image processing. Text detection has a wide range of uses. It assists visually disabled people in "reading" street signs and currency [423, 424]. Identifying and recognizing house numbers and street signs in geographic information systems makes it easier to create digital maps [425, 426]; for further details, please see Table 2. Different types of problems (different fonts and languages, text rotation and perspective distortion, densely arranged text localization, Broken and blurred characters) are considered the main research problems in this field.

### 1.5.4. *Traffic Sign and Traffic Light Detection (TSTLD)*

The automatic detection of traffic signs and traffic lights has gotten a lot of attention due to self-driving technology advancements. While the computer vision community has primarily focused on identifying general objects rather than fixed patterns such as traffic lights and traffic signs in recent decades, it would be wrong if assumed that traffic light recognition is not complex (Table 2). Traffic signs also have different research problems (illumination changes, motion blur, bad weather, real-time detection). Generally, tracking by detection and detection by tracking gained more attention than visualizing static lights.

### 1.5.5. *Remote Sensing Target Detection (RSTD)*

The techniques of remote sensing have given people new opportunities to investigate inside the planet. As remote sensing images have increased in recent years, remote sensing target detection (detecting an airplane, ship, oil-pot, etc.) has become a research hotspot. Target detection via remote sensing has numerous applications, including military investigation, disaster relief, and urban traffic management. In RSTD, some difficulties (detection in “big data,” occluded targets, domain adaptation) have attracted different solutions, the short analysis presented in Table 2.

## 1.6. Challenges/problems for tiny Object detection & Deep Domain adaptation

### 1.6.1. ***Problem 1****: Insufficient information in Individual feature layers for tiny objects detection/domain adaptation*

Due to pooling and subsampling processes, deep CNN architectures generate hierarchy feature maps. As a result, different layers of feature maps with various spatial resolutions are produced. It is a well-known phenomenon that the early-layer feature maps have a better resolution and represent smaller reception fields. Simultaneously, they lack high-level semantic information that is critical for object detection. On the other hand, the latter-layer feature maps contain more semantic information, which is necessary for identifying and classifying objects, such as distinct object positions or illuminations. Higher-level feature maps help identify large objects but may not detect smaller objects and their domain. The latter feature maps lose spatial information after downsampling numerous times in deep CNN systems. In earlier (or shallower) feature maps, a tiny object of 32x32 pixels is visible, but not in later (or deeper) feature maps. As a result, detecting small objects and domain adaptation using only low-level or high-level feature map characteristics is considered insufficient.

**Solution: Combining features from both shallow and deep layers.**

Several deep CNN-based approaches combine lower-level and higher-level feature maps to provide the spatial and semantic information necessary to recognize tiny objects. Two ways can accomplish feature map fusion.

**1) Bottom-Up Approach:** The conventional feedforward CNN design incorporates this method. After pooling procedures, feature maps shrink from the first to the last layers. Several bottom-up feature maps are directly combined in the final detection layers.

**2)Top-Down Approach:** This approach can be thought of as an attention mechanism that sends higher-level semantic information to lower-level feature maps. To increase the spatial resolution of the feature maps, it commonly employs a convolution-deconvolution or encoder-decoder network with an upsampling operation in the decoder. Furthermore, the skip paradigm or lateral connection is frequently utilized to connect lower-layer feature maps to higher-layer feature maps while avoiding intermediary layers. Detection layers use fused feature maps. Summation, production, concatenation, and global pooling are standard methods used for combining feature maps.

### 1.6.2. ***Problem 2****: Insufficient Context Information for tiny objects detection/domain adaptation*

Tiny objects usually have lower resolutions, which makes it difficult to distinguish them from other items. Contextual information is crucial in tiny object detection because tiny objects carry limited information [262-265]. From a "global" image level to a "local" image level, contextual information has a vital role in object recognition. A global image-level considers image statistics from the entire image, whereas a local image level considers contextual information from the objects' surrounding areas. Three types of context features can be identified [264]:

**1)Local pixel context information:** The patches of pixels surrounding an item, including edges, colors, textures, and so on. The size of the detection window in object detection networks can be increased to collect local pixel context.

**2)Semantic context information:** The likelihood that an object would be recognized in some nearby situations, such as events, activities, or scene categories.

**3) Spatial context information:** The spatial positioning of other objects in the image, such as the probability of locating an object in certain positions relative to other objects in the image. The subject's shoulder and neck, for instance, are constantly close to their face in face detection systems.

**Solution: Contextual information is incorporated into the detection network.**

The local pixel context is frequently provided by extending filter sizes to collect extra information around the objects. Deeper features from images, such as deconvolution layers or recurrent neural networks, are commonly used to add semantic context (RNNs).

### 1.6.3. ***Problem 3****:Tiny object class imbalance for tiny objects detection/domain adaptation*

The term "class imbalance" refers to the unequal distribution of data between classes. In other words, it is also considered the imbalance of information distribution between foreground and background instances. By densely scanning the entire image, area proposal networks are employed in object detection to create candidate regions containing objects. The anchors are rectangular boxes that have been extensively tiled throughout the full input image. Anchor scales and ratios are pre-determined based on the sizes of target objects in the training dataset. When detecting tiny objects, the number of anchors generated per image is higher than when recognizing large objects. Positive instances are only those anchors with a high Intersection over Union (IoU) with the ground truth bounding boxes. Because most anchors have little or no overlap with the ground truth bounding boxes, they are regarded as negative examples. Positive examples are a minuscule percentage when densely generated anchors are matched with sparsely located real objects in the images, resulting in a high-class imbalance, such as a class ratio of 100:1 to 1000:1. The anchor-based object detection method has several flaws. First, because of the scarcity of ground-truth bounding boxes and the IoU matching methods between ground-truth and anchors, negative examples vastly outnumber positive ones, resulting in models that favor the negative class.

Second, the dense sliding window technique has a high temporal complexity (O(h2w2), where h is the height of the anchors and w is their width, making training slow.

**Solution: Balancing positive/negative examples during training**

There are two main approaches used for that purpose,

1) **Data based approach**
2) **Loss function-based approach**

The data-driven technique is to modify the foreground and background example numbers to equalize the weights of positive and negative instances roughly. There are two standard sampling methods: complex sampling and soft sampling. Soft sampling assigns different weights to examples, whereas complex sampling selects a subset of samples. Random sampling, for example, is frequently used to select examples at random to fulfill a specific ratio. Another option is to sample more of the problematic examples with high losses. For instance, a machine learning model may be taught first, and then the false positives could be deemed complex examples in the second round of training. (Shrivastava et al., 2016) presented the Online Hard Example Mining (OHEM) approach, which conducts a single forward pass on the computed region of interest (RoI) and calculates losses for all ROIs [266]. Then, based on their loss function values, examples are ranked, and the ones with the most significant loss are chosen to be used in the next round of training because the present trained network model performs the worst on them. To sample more training examples from tough instances, (Pang, Chen, et al., 2019) suggested an IoU-balanced sampling technique [267-268]. In terms of soft sampling, (Cao et al., 2020) proposed a method for selecting samples based on their importance, in which the volume of positive examples is measured by their IoU scores with the ground truth bounding boxes. The importance of negative examples is calculated by considering both local and global region properties [269]. For imbalanced classes, loss function-based algorithms help during balancing background and foreground weights. For example, AP loss (Chen et al., 2019) re-weights samples using an average precision loss [268]. DR Loss (Qian et al., 2019) reweights samples depending on the distribution of foreground examples compared to background examples [270].

#### *1.6.4. **Problem 4**: Shortage of tiny objects examples for tiny objects detection/domain adaptation*

Network models trained on various scales usually show better results with larger objects but poorly with tiny objects. A lack of small-scale anchor boxes to match the tiny objects and insufficient samples to be appropriately compared to the ground truth could be among the reasons. The anchors are feature mappings from specific intermediate layers in a deep neural network projected back to the original image. Tiny objects have a difficult time generating anchors. Furthermore, the anchors must be matched to the bounding boxes of the ground truth. The following is an example of a popular matching method: An anchor is designated as a positive example if it has a high IoU score about a ground truth bounding box, such as more than 0.9. In addition, each ground truth box's anchor with the most excellent IoU score is identified as a positive example. As a result, small objects typically have a limited number of anchors that match the ground truth bounding boxes, i.e., a limited number of positive examples.

**Solution: Use methods that can generate/match more small-object anchors.**

Following existing techniques help to solve the problem.

**1)multi-scale mechanism.**

Anchors of various scales can be generated using multi-scale topologies with independent branches for tiny, medium, and large-scale objects.

**2)Matching strategy.**

Setting anchor scales and ratios adaptively to assist more anchors in matching to small object ground truths.

**3)Increasing positive examples of small objects.**

Allowing anchors to overlap at the region proposal stage will generate more anchors.

## 2 OBJECT DETECTION TRENDING APPROACHES

### 2.1. General techniques to solve tiny object detection/DDA-OD problems

Object detection (OD) and tracking have sparked a lot of interest due to the production of high-powered computers, the availability of high-quality and inexpensive video cameras, and the need for automated video analysis. Detecting and tracking moving objects, such as people and cars, plays an essential role in video data tracking through video frames. Computer vision applications, such as video surveillance, person tracking, traffic control, semantic annotation of images, detecting moving objects and static image objects, and finding objects in video streams can be considered the first relevant data extraction stage. Before moving on to object detection methods/models/datasets, some critical approaches need to know before further analysis of object detection (OD). To overcome different problems related to OD that are presented in Table 2, different strategies and methods proposed by various researchers, few of them are as follow:

#### *2.1.1. Face detection-based research (Trending issue for tiny object objects and domain adaptation)*

Face recognition has been an extensively researched area and gets a lot of success. Face recognition differs from general object detection because faces include distinguishing facial characteristics, such as the nose, eyes, mouth, and relative positions related to one another. However, when face sizes are tiny, such as less than 16x16 pixels, and features are not identifiable, it remains challenging. For small face detection, a variety of approaches have been developed. Table 6 shows the state-of-the-art algorithms. The state-of-art algorithms are shown in Table 6.

**Technique#1:** *Feature maps improvements for small faces/DDA-OD*

This technique is used for skip connections to integrate lower, intermediate, and higher-layer features. In general, this method helps in merging several feature maps. (Tian et al., 2018) offered an iterative feature map generation approach. All feature maps from the backbone network were fed back to the beginning of the network to extract more semantic information for small items [271]. (Samangouei et al., 2018) used an ROI-based block normalization layer to feed the combined lower and higher layer features [272]. (Tian et al., 2018) used skip connections to merge four feature maps into four new feature maps for detection [271]. By element multiplica-tion, the input features were fused with the next level characteristics. In addition, the fused features were combined with the original input features to create the final features, which showed to be effective in detecting tough little faces. (Zhu et al., 2017) merged lower and higher-level features by downsampling lower-level features to the size of higher-level features and then concatenating them using L2 normalization [273]. (Luo et al., 2019) used a bilinear upsampling technique to blend lower-level data with surrounding features [274]. By recurrently crossing the network [275], (Yoo et al., 2019) designed a feature map creation approach by recurrently passing the network.

**Technique#2:** *Incorporate context information of small faces(a problem in detection based domain)*

(Bai et al., 2018) shown that adding context information to face detection improves performance considerably for small faces. However, too much background information can hinder performance on small faces [567]. By extending the receptive areas surrounding faces, several approaches incorporated context information. (Najibi et al., 2017) used higher filter sizes in their convolutional networks, such as 5x5 and 7x7 filters instead of 3x3 filters [276]. An agglomeration connection module (Wang et al., 2017) used the feature fusion approach to integrate lower-level features with higher-level features. Lower feature maps travel through an Inception-like network in this module to boost semantic information, and subsequently, lower feature maps are concatenated with higher feature maps [277]. Around each bounding box, (Samangouei et al., 2018) inserted context information [278]. Extra context information from bodies and shoulders was combined (Tang et al., 2018), and a semi-supervised approach was utilized to produce labels for other body parts [279]. (Tian et al., 2018) used a segmentation branch to offer further context and semantic information without adding new annotations. The segmentation branch shared the same receptive field of detection, making it an additional source for more discriminative features [280]. To incorporate retrieved-context characteristics, (Li et al., 2019) employed the dense block structure (Huang et al., 2017). To eliminate false positives, (Zhu et al., 2017) integrated body information. Additional RoI-pooling techniques were used to acquire the body features to expand the receptive fields. Both classification and bounding box regression [566] [281,282] used a combination of facial and body features.

**Technique#3:** *Correcting foreground/background class imbalance for small faces(Have a vital role in detection and domain adaptation)*
detector networks for small face detection typically place a large number of small anchors on the images, resulting in a large number of negative anchors and a small number of positive anchors, resulting in a high false-positive rate. These three basic techniques are used to deal with the class imbalance object detection problem.

***a) Anchors for filtering***. (Zhang et al., 2017) employed a max-out background label, which predicted many background label scores and chose the highest as the final score [283]. (Chi et al., 2019) employed two-step classification on the lower layers to filter out false positives for small faces, which improved the classification results by balancing positive and negative samples.

***b) Sampling***. (Zhang et al., 2017) used hard-negative mining to reduce the negative-to-positive ratio to no more than 3:1 [283]. Hard-negatives mining was also employed by (Najibi et al., 2019a). If the overlap with the ground-truth bounding box was more significant than 0.5, the anchor was labeled positive. (Li et al., 2019) proposed a balanced-data-anchor-sampling technique [284] for selecting large and small size anchors with equal probability. To lower the false positive rate of small items, (Tang et al., 2018) presented a Pyramid box that used the max-in-out technique on both positive and negative samples [285]. (Wang, Li, Ji, & Wang, 2017) used online complex example mining (OHEM) to sort the examples by loss and choose the top examples with the most significant loss as complex examples. They also employed a 1:1 ratio for positive and negative complex examples in each mini-batch [286].

**c)Training on multiple scales.** (Wang et al., 2017) scaled input images to produce objects of varying sizes and small objects can be resized to larger objects to match more anchor boxes [287]. (Najibi et al., 2017) developed a multi-scale network with three separate convolutional branches to recognize different face scales: tiny, medium, and large faces [284]. An element sum feature from conv4 and conv5 was used to identify small faces. Conv5 was used to detect medium faces directly. After conv5, large faces were recognized using max pooling. The input photos were scaled to ratios of 0.5, 1, and 2 of the original resolution (Hu et al., 2017). Then, to capture different scales of faces, two types of feature maps were used [288].

### *2.1.2. Generic object detection research (Deep learning Technique to solve tiny object detection problems)*

The deep learning approaches established in generic object detection research effective for tiny object detection are summarized in this section.

**Technique#1:** *Feature map improvement in tiny objects(for domain adaptation and detection, feature maps perform a significant role)*
Localization relies on low-level features, whereas classification relies on high-level features. Several deep neural networks have improved their performance for tiny object detection by combining low-level, and high-level feature maps at the detection layers. One group of network topologies integrated feature maps from multiple layers using a bottom-up approach (Fu et al., 2017; Lin et al., 2017; Kong et al., 2017) [289,290][506]. Considering both global and local information, (Kong et al.,2018) presented a non-linear feature map transformation. The non-linear transformation parameters could be learned and shared between layers. After the transformation, each transformed layer produced detection results applied to feature maps in separate layers [290]. In an "encoder-decoder" design (Yang et al., 2019), deconvolutional layers were coupled with feature maps from convolutional layers in an "encoder-decoder" architecture [507]. Skip connections were utilized by (Bell et al., 2016) to add lower-level feature maps to higher-level feature maps [508]. Convolutional layers with varying receptive fields were used to create features (conv3, conv4, and conv5). These characteristics were concatenated and standardized before being supplied into detection modules. The feature pyramid fusion process was used at two levels (Pang et al., 2019) [509]. It built an image pyramid for global data and merged features from four levels of the image pyramid with the original features from the SSD architecture. Features from both the previous and current layers were combined to create local spatial information.

**Technique#2:** *Incorporate context information for tiny objects(context information also necessary for detection)*
Small object context information is divided into two categories: local contact and sematic context information. Larger bounding boxes and proposal boxes could be used to add more local context information in deep neural networks. (Cai et al., 2016) used bounding boxes 1.5 times the size of object regions to offer extra local context, which was effective for including more of the surrounds of small objects. For detection layers, the added context information was coupled with object features [510]. (Zagoruyko et al., 2016) used four scales to incorporate local context information into area proposal boxes: 1x, 1.5x, 2x, and 4x. Before being fed into the detection and classification layers, the outputs from four regions were pooled and concatenated using ROI-pooling [511]. (Fu et al., 2017) used deconvolutional layers with "skip connections" to include additional semantic context information, resulting in superior detection results on small objects. Rather than stacking deconvolutional layers on top of convolutional layers, the deconvolution layers were intended to be considerably shallower than the convolutional layers, and an element-wise product was utilized [289]. Deconvolutional layers were also employed by (Cai et al., 2016) to improve the resolutions of feature maps [510]. To collect global information of the input image, (Bell et al., 2016) used four Recurrent Neural Networks (RNNs) [508]. RNNs added semantic context information to the objects, and 1x1 convolutions integrated all of the data. Furthermore, (Zhang et al., 2018) [512] passed higher-level feature maps to lower-level features.

**Technique#3:** *Small objects foreground/background class imbalance(challenge for tiny object detection based domain adaptation)*
A data-based approach and a loss function-based approach are two techniques for addressing the foreground and background class imbalance. (Cai et al., 2016) used bootstrap sampling to sample negative examples depending on their loss values in their data-based technique to address the class imbalance [510]. To balance the foreground and background examples, (Zhang et al., Wen et al., 2018) employed a two-step regression technique. Certain easy negatives were omitted to make the ratio between positives and negatives nearly even [512][548]. Before filtering out the bounding boxes lacking objects, (Kong et al., 2017) implemented an abjectness [506]. (Galleguillos et al., 2010) utilized a loss function that gave hard-negative examples more weight in the loss function-based method [549].

**Technique#4:** *Training examples increase for tiny objects*
To enhance the number of training examples for tiny objects, many strategies have been devised. They include multi-scale learning neural network architectures, scale transformation, and adaptive anchor box matching. Several neural network architectures have been devised for multi-scale learning to solve insufficient instances of small things in training classifiers, i.e., training detector networks for objects of various sizes. Due to the increased quantity and variety of training examples, (Singh et al., 2018) demonstrated the effec-tiveness of a training strategy that used objects of diverse scales and positions to boost the detection performance on small objects [550]. For detection, small objects were up-sampled and fed into convolutional networks. During training, only the layers whose feature maps contained target items within a given range were activated, allowing small objects and medium and enormous objects to be trained. Machine learning has been used to determine perfect scales and ratios for anchor boxes and do adaptive matching of anchor boxes instead of pre-defined anchor boxes (Redmon et al., 2017). Ground truth bounding boxes, for example, can be sorted into clusters based on their scales, and anchor boxes can then be matched to ground-truth bounding boxes with similar scales [551].

### *2.1.3. In aerial images, object detection*

There are four types of approaches for recognizing objects in aerial images: i) template matching-based, (ii) knowledge-based, (iii) OBIA-based, and (iv) machine learning-based (Cheng et al., 2016) [552]. Deep learning-based algorithms have outperformed all others in recent years. CNN's were commonly fine-tuned on aerial images after being pre-trained on massive image datasets like the ImageNet and COCO datasets. To improve performance, new deep neural networks were developed for the unique attributes of objects in aerial images, such as multi-scale and multi-angle. For example, to get good performance on remote sensing images, (Dong et al., 2018) suggested rotation-invariant models [553]. Furthermore, weakly supervised learning approaches (Peng et al., 2018) have been proposed to learn high-level features in an unsupervised manner to capture object structure information in remote sensor images [554].

**Technique 1***: Object orientation handling in aerial images.*

Objects in aerial images can be rotated or oriented in different ways.

To overcome this problem, deep neural network-based detectors have been developed. The Rotation-Invariant CNN (RICNN) is a novel rotation-invariant layer in the standard CNN architecture (Cheng et al., 2016) [555]. In (Cheng et al., 2018), the Rotation-Invariant and Fisher Discriminative CNN (RIFD) was proposed to include a rotation-invariant regularizer and a fisher discrimination regularizer on multi-scale features from CNN[556]. The rotation-invariant regularizer constrained the CNN features to be similar for within-class examples but dissimilar for different classes. In contrast, the fisher discrimination regularizer denied the CNN features identical for within-class examples but distinct for other classes. Anchor rotation algorithms have recently been proposed by (Yang et al., 2018) to achieve rotation invariance in one-stage object detectors. To increase detection performance on aerial images, a feature refinement technique was presented. The position information of bounding boxes was encoded to the associated feature points using feature interpolation to improve feature reconstruction and alignment. R-Net presented a network to produce rotatable region ideas (Yang et al., 2018) [557].

**Technique 2***: Take into account the context of aerial image objects.*

More context information for small objects has been added in the detection networks through merged feature maps and dilated convolutions. Multiple convolutional layers' feature maps can be combined to create a new feature map. To boost performance on small-scale object detection, CNN models can be enhanced with dilated convolutions.

**Technique 3:** *For aerial image objects, correcting foreground and background class imbalance.*

IoU-Adaptive Deformable R-CNN was presented in (Yan et al., 2019) to address the class imbalance issue in training classifiers in object detectors [558]. It is based on Faster RCNN. An IoU-guided detection framework was presented to limit the loss of tiny object information during training by evaluating the numerous roles that IoU can play in different areas of network models. In addition, to increase detection accuracy, an IoU-based weighted loss was devised to learn the IoU information of positive ROIs. Finally, the class aspect ratio constrained non-maximum suppression (CARC-NMS) method was devised to improve detection precision.

**Technique 4:** *Increase the number of aerial picture object training examples.*

Multi-Scale and Rotation-Insensitive Convolutional Channel Features (MsRi-CCF) combines robust low-level feature generation, classifier development with outlier reduction, and detection with a power law for geospatial object detection by (Wu et al., 2018) [559].

### *2.1.4. Instance segmentation approach for small object detection*

For instance, segmentation deep CNNs have also been employed to object detection, in contrast to the popular bounding-box-based object detectors mentioned in the previous sections. The main disadvantages of segmentation algorithms are the time-consuming pixel-by-pixel labelling and the computation and memory requirements. Each pixel of an object is vital for small object detection, and employing pixel information could produce good results. One of the earliest ways to use CNNs for semantic segmentation was FCN (Long et al., 2015). FCN uses CNNs that don't have fully connected layers, allowing the input image to be any size. It takes advantage of pooling layers to speed up computation and expand the reception field [560]. To address finding sufficient pooling layers, U-Net (Ronneberger et al., 2015) was presented using an encoder-decoder design based on FCN [561]. It uses a U-shape design to strike a balance between localization accuracy and context information efficiency. Pooling layers are used in the encoder to reduce the layer size progressively, while up-convolution is used in the decoder to raise the layer size gradually.

Furthermore, U-Net employs short-cut connections from the encoder to the decoder to aid the decoder in recovering fine-grain data. Large reception fields result in lesser localization accuracy regarding the trade-off between reception field and localization accuracy. However, due to a lack of background information, localization accuracy may suffer when the reception field is too small.

FPNs (Feature Pyramid Networks) is a hybrid of FCN and Faster R-CNN. FPNs added a third output, instance mask prediction for segmentation, to the two predictions Faster R-CNN generates: I bounding box localization and (ii) bounding box recognition. FPNs also applied various novel techniques for additional improvements, such as new ROI align layers, multitask training, and better backbone networks.

Other strategies based on segmentation methods have also been presented for small item detection. Capsule networks with deconvolutional capsules were proposed to expand the primary layers in network topologies to accommodate more context in- formation (LaLonde et al., 2018) [562]. Segmentations could be fine-tuned by combining features from different layers using bottom-up and top-down network architectures (Ronneberger et al., 2015) or by employing pyramid pooling layers to segment objects at several scales as in DeepLab (Chen et al., 2018) [563]. To improve the number of training instances for small objects, a more robust embedding might be generated by combining features from several models and employing unsupervised and supervised learning together to form a multi-scale representation (Lin, Milan et al., 2017) [565]. Concept Mask used a semi-supervised learning strategy to train a deep neural network with image-level la- bels in (Wang Lin et l., 2018). The findings were then modified and expanded to predict attention maps [564]. Finally, a class segmentation network based on attention was trained.

## 2.2. Deep Domain Adaptive Object Detection (DDAOD)

Object detection is an essential and challenging process in computer vision used in several applications, i.e., autonomous driving, robot vision, and human-computer interaction. Most state-of-the-art DL-based object detection methods presume that training and test data come from the same distribution source. These detection models require a significant number of practical training samples. In practice, gathering advanced annotated data is a time-consuming and costly process. Deep domain adaptation is a new learning paradigm that addresses the challenges described above.
Deep domain adaptation (DDA) in some computer vision tasks such as image classification and semantic segmentation [89, 90] is gaining great success. It is predicted that object detection accuracy can be boosted using DDA. Due to detection field advancements, much experimental work was done to experience the DDA role in object detection, and many DDAOD methods have been proposed in recent years. Some studies focused on review domain adaptation [89,90] and deep learning-based object detection [91], but not many comparative research contributions on DDAOD published yet. As a result, we highlighted this domain in the OD survey for an entire field understanding. This approach discussion aims to examine the current state of DDAOD methods and provide some insight into potential research trends.
In this section, we first explained several factors that will be helpful later to categorize DDAOD methods and then reviewed related DDAOD methods.
Domain shift can be addressed using five different mechanisms and categories of DDAOD: discrepancy-based, adversarial-based, reconstruction-based, hybrid, and others.

- *One-step vs. multi-step adaptation methods:* Weather the source and target domains are closely linked, information transfer may be completed in a single step. Although there is little overlap between the two parts, multi-step DA uses a series of intermediate bridges to link two different domains and then execute one-step DA using this bridge.
- *Labeled data from the target domain:* DDAOD may be classified as supervised, semi-supervised, weakly-supervised, few-shot, or unsupervised based on labeled data and domain.
- *Base detector:* Domain adaptive detection methods are often based on existing excellent detection models such as Faster RCNN, YOLO, SSD, and others.
- *Is the method's source code open source or not?* This aspect shows whether the method's source code is available on the internet. The relation will be given if it is open source.

We classify DDAOD methods in Table 3. based on the categorization as mentioned above factors and then review them in the following subsections.

### 2.2.1. Discrepancy-based DDAOD

Some discoveries about deep learning-based detection highlighted that the deep network-based detection models with labeled or unlabeled target data could be considered a research problem. The discrepancy-based DDAOD approaches can reduce changes in the domain. For domain adaptive object detection, Khodabandeh et al. [92] proposed a robust learning approach. The authors defined the problem as "training with noisy labels." The final detection model is trained using a collection of noisy objects bounding boxes obtained from a detection model trained only in the source domain [93][94] Table 3.1.

Table 3.1 *Discrepancy-based DDAOD and obtained results with data analysis*

| Method applied | Authors | Used Dataset | Results(MP[2]) |
|---|---|---|---|
| ***Faster RCNN-Unsupervised-One step Domain Adaptive*** | (Khodabandeh et.al,2019) [92] | Cityscapes [119] Foggy[120] | **car AP** :36.5, oracle: 43.5 |
| | | Cityscapes KITTI [121] | **car AP: 77.6**, oracle: 90.1 |
| | | KITTI Cityscapes | car AP: 43.0, oracle: 68.1 |
| | | SIM 10k [122] Cityscapes | car AP: 42.6, oracle [3]: 68.1 |
| | (Cai et.al, 2019 ) [93] | Cityscapes Foggy | 35.1 |
| | | SIM 10k Cityscapes | car AP: 46.6 |
| | | Synthetic [123], COCO [124] | 20.7 |
| | (Cao et.al,2019) [94] | Synthetic YTBB [125] | 22.8 |
| | | Caltech [126] visible KAIST | F1 of annotation: 0.75 |
| | | multispectral [127] | Miss rate:32.66 |

### 2.2.2. Adversarial-based DDAOD

Domain discriminators and adversarial preparation are used in adversarial-based DDAOD approaches to promote domain confusion between the source and target domains. Domain discriminators distinguish between data points taken from the source and goal domains. Adaptive to the environment, the first paper to address the domain adaptation issue for object detection is Faster RCNN [95]. The authors used H-divergence to assess the divergence between the data distributions of the source, target domains and adversarial function training. Image-level adaptation, instance-level adaptation, and consistency review are the three adaptation components [96-103]—Table 3.2.

### 2.2.3. Reconstruction-based DDAOD

Reconstruction-based DDAOD assumes that reconstructing the source or target samples which helps in the efficiency of domain adaptation object detection. Arruda et al. [104] suggested an unsupervised image-to-image translation system for cross-domain car detection. CycleGAN was investigated to see whether it could create a false dataset by converting images from daytime to nighttime. The final detection model is trained on a fictitious dataset using annotations from the source domain [105-108]-Table 3.3.

Table 3.2 *Adversarial-based DDAOD and obtained results with data analysis*

| Method applied | Authors | Used Datasets | Results mAP[2] (%) |
|---|---|---|---|
| ***Faster RCNN-Unsupervised-One step Domain Adaptive*** | Chen et.al. [95], 2018 | Cityscapes   Foggy | 27.6 |
| | | Cityscapes   KITTI | car AP: 64.1 |
| | | KITTI   Cityscapes | car AP: 38.5 |
| | | SIM 10k   Cityscapes | car AP: 39.0 |
| | Zhu et.al. [96], 2019 | Cityscapes   Foggy | 33.8 |
| | | KITTI   Cityscapes | car AP: 42.5 |
| | | SIM 10k   Cityscapes | car AP: 43.0 |
| | Wang et.al. [97][4], 2019 | Cityscapes   Foggy | 31.3 |
| | | Cityscapes   Udacity | 48.5 |
| | | SIM 10k   Cityscapes | car AP: 41.2 |
| | | SIM10K   Udacity | car AP: 40.5 |
| | | Udacity [128]   Cityscapes | 50.2 |
| | Saito et.al. [98], 2019 | Cityscapes   Foggy | 34.3 |
| | | PASCAL [129]   Clipart [130] | 38.1 |
| | | PASCAL   Watercolor [109] | **53.3** |
| | | SIM 10k   Cityscapes | **car AP:47.7** |
| | He et.al. [99], 2019 | Cityscapes   Foggy | 34.0 |
| | | Cityscapes   KITTI | car AP:72.1 |
| | | KITTI   Cityscapes | car AP:41.0 |
| | | SIM 10k   Cityscapes | car AP: 41.1 |
| | Shen et.al [100], 2019 | Cityscapes   Foggy | 37.9 |
| | | Cityscapes   KITTI | **car AP:72.7** |
| | | KITTI   Cityscapes | 41.9 |
| | | PASCAL   Clipart | 41.5 |
| | | PASCAL   Watercolor | **55.2** |
| | | SIM 10k   Cityscapes | car AP: 42.6 |
| | Zhang et.al [101], 2019 | SYNTHIA [130]   Cityscapes | 33.2 |
| | | VKITTI [131]   Cityscapes | car AP: 52.8 |
| | Zhuang et.al. [102], 2020 | Cityscapes   Foggy | 36.2 |
| | | SIM 10k   Cityscapes | car AP:47.1 |
| | Chen et.al. [103], 2020 | Cityscapes   Foggy | 39.8, oracle: 40.3 |
| | | PASCAL   Clipart1k | 40.3 |
| | | Sim10k   Cityscapes | 42.5 |
| Mask-RCNN,*Unsupervised* | Zhang et.al [101], 2019 *One step Domain Adaptive* | SYNTHIA [130]   Cityscapes | 32.2 |

Table 3.3 *Reconstruction-based DDAODand obtained results with data analysis*

| Authors | One/Multi DA/ Label Target | Basic Detector | Dataset and Task | Results-mAP[2] (%) |
|---|---|---|---|---|
| Arruda et.al. [104], 2019 | One-step Domain adaptive *Unsupervised* | Faster R-CNN, Replaceable | BDD100k [132], day-night | **86.6±0.7**, oracle: 92.0±0.8 |
| Lin et.al [105],2019 | | YOLO / Faster R-CNN | BDD100k, day-night | mAP of YOLO / Faster R-CNN: **41.9/67.0** |
| Guo et.al. [106], 2019 | | Faster RCNN, replaceable | KAIST visible   thermal | Miss rate: 42.65 |
| Devaguptapu et.al. [107],2019 | | Faster RCNN | FLIR ADAS [133], visible thermal<br>KAIST visible   thermal | 61.54<br>53.56 |
| Liu et.al. [108],2019 | | Faster RCNN, replaceable | SENSIAC [134], visual middle-wave infrared | 91.7 |

### *2.2.4. Hybrid DDAOD*

To improve efficiency, hybrid DDAOD offered more than two mentioned mechanisms simultaneously. Cross-domain weakly controlled object detection is a novel activity proposed by Inoue et al. [109] in which image-level annotation is accessible in the target domain. A two-step progressive domain adaptation technique is proposed to address this challenge. On two forms of artificially and automatically produced samples, this method fine-tunes the detector. A CycleGAN-based image-image translation is used to artificially generate models while automatically delivered samples obtained by pseudo-labelling techniques [110-116] Table 3.4.

### *2.2.5. Other DDAOD*

Other DDAOD methods do not fit into any of the four categories mentioned above. They attempt domain alignment using other mechanisms such as graph-induced prototype alignment [117] and categorical regularization [118]. Xu et al. [117] propose the graph-induced prototype alignment (GPA) system and integrate it into a two-stage detector, i.e. Faster R-CNN. It addresses source issues and target domains on a local instance level and class-imbalance in cross-domain detection tasks. The GPA system outperforms current approaches by wider range standards, according to experimental findings.

Xu et al. [118] propose a categorical regularization method because previous work has failed to align critical image regions and essential instances across domains. It can be used in several Domain Adaptive Faster R-CNN methods as a plug-and-play component. There are two regularization modules planned. The first module takes advantage of classification CNNs' insufficient localization capacity, while the second takes advantage of categorical consistency between image-level and instance-level predictions Table 3.5.

Table 3.4 *Hybrid DDAODand obtained results with data analysis*

| Authors | One/Multi DA | Label Target | Basic Detector | Dataset [1]& task | Results mAP[2] (%) |
|---|---|---|---|---|---|
| Inoue et.al.<br>[109], 2018 | Multi-step<br>DA | Weakly-<br>Supervised | SSD,<br>replaceable | PASCAL Clipart1k<br>Watercolor2k | **46.0**, Ideal case: 55.4<br>**54.3**, Ideal case: 58.4 |
| Shan et.al.<br>[110], 2019 | Multi-step<br>DA | Unsupervised | Faster RCNN | Comic2k<br>Cityscapes Foggy<br>Cityscapes KITTI<br>KITTI Cityscapes<br>KITTI VKITTI-Rainy<br>Sim10k Cityscapes<br>Sim10k KITTI | **37.2**, Ideal case: 46.4<br>mAP: 28.9<br>car AP: 65.6<br>car AP: 41.8<br>mAP: 52.2<br>car AP: 39.6<br>car AP: 59.3 |
| Kim et.al.<br>[111], 2019 | Multi-step<br>DA | Unsupervised | Faster RCNN | Cityscapes Foggy<br>PASCAL Clipart1k<br>Watercolor2k<br>Comic2k | 34.6<br>**41.8**<br>52.0<br>**34.5** |
| Kim et.al.<br>[112], 2019 | One-step<br>DA | Unsupervised | SSD | PASCAL Clipart1k<br>Watercolor2k<br>Comic2k | 35.7<br>49.9<br>26.8 |
| Rodriguez<br>et.al. [113],<br>2019 | Multi-step<br>DA | Unsupervised | SSD | Cityscapes Foggy<br>PASCAL Clipart1k<br>Watercolor2k, Comic2k<br>Sim10k Cityscapes | 29.7<br>44.8<br>57.3, 39.4<br>car AP:44.2 |
| Hsu et.al.<br>[114], 2020 | Multi-step<br>DA | Unsupervised | Faster RCNN | Cityscapes BDD100k<br>Cityscapes Foggy<br>KITTI Cityscapes | 24.3, oracle: 43.3<br>36.9, oracle: 39.2<br>car AP: 43.9 oracle: 55.8 |
| Yu et.al. [115],<br>2019 | Multi-step<br>DA | Unsupervised | Faster RCNN | Cityscapes Foggy<br>KITTI Cityscapes<br>Sim10k Cityscapes | 38.2,oracle:42.5<br>Car AP:46.4, oracle:62.7<br>car AP:52.3, oracle: 62.7 |
| Zheng et.al.<br>[116] . 2020 | One-step<br>DA | Unsupervised | Faster RCNN | Cityscapes Foggy<br>Cityscapes KITTI<br>Sim10k Cityscapes | 38.6, oracle:43.3<br>car AP: 73.6, oracle:88.4<br>(AP: 41.0, oracle:85.4)<br>car AP: 43.8, oracle:59.9 |

Table 3.5 *Other DDAODand obtained results with data analysis*

| Authors | One/Multi DA | Label Target | Basic Detector | Dataset [1]& task | Results mAP[2] (%) |
|---|---|---|---|---|---|
| Xu et.al. [117],<br>2020 | One-step<br>DA | Unsupervised | Faster RCNN | Cityscapes Foggy<br>KITTI Cityscapes<br>Sim10k Cityscapes | 39.5<br>Car AP: 47.9<br>Car AP: 47.6 |
| Xu et.al, [118],<br>2020 | One-step<br>DA | Unsupervised | Faster RCNN | Cityscapes BDD100k<br>Cityscapes Foggy<br>PASCAL Clipart1k | 26.9, oracle:38.6<br>37.4, oracle:42.4<br>38.3 |

**Table note: 1.** Dataset's source and other literature state-of-the-art can be found in reference. 2 Bolded red, green, and blue highlights the first place, second place and third place, respectively.3. Oracle represents Faster R-CNN trained on the target domain. 4. Results of UDA are included in this table.

## 2.3. Object detection trending approaches for tiny objects

Tiny objects can be defined in one of two ways. In the real world, one applies to smaller objects in dimension and two applies to large objects. In the MS-COCO [141] metric evaluation, another description of tiny objects is stated. Objects with areas smaller than or equivalent to 32 x 32 pixels are classified as "tiny objects. This size threshold is widely used throughout the principles for datasets about popular objects. In Fig 1, we presented a few examples of tiny objects (e.g., "baseball," "tennis," "badminton shuttle," "cricket ball," insects in air and "pg" on a street sign). Many object detectors are good at detecting medium and large objects but not so good at detecting tiny objects. Mainly there are three challenges in detecting tiny objects. Tiny objects have fewer visual cues required to differentiate them from the context or other related categories. There are a lot of options for tiny object placement in a big image.

General object detection is always a more attractive field for researchers. Typically, large object detection due to autonomous system trends and driverless vehicle reputation has become part of their target consideration. We tried to elaborate on general object detection and tiny object detection and tried to analyze problems related to them. In the field of computer vision, many models, methods, and algorithms helping to solve the detection problems for tiny objects inside the image and video data, i.e., multi-scale feature learning, data augmentation, training strategy context-based detection, etc. GAN-based detection.

In Table 4, we presented a historical outlook of different object detection models with the help of recent and old research contributions. Instead of giving a detailed description, we tried to highlight an overview of the field. Most of the results are presented in tabular form to get sight in the entire literature quickly. In the last column of this Table 4, we presented methods taxonomy step by step from left to right. It means which way invented at which duration and from method-to-method version support in object detection. In Table 5, we displayed a research scenario about tiny object detection approaches—literature cited inside this table concerning each small object detection approach with a proper understanding of related methods. For easy understanding, data is divided into categories and subcategories and mentioned detection techniques and methods inside. Some strategies for tiny object detection and found results are presented in Table 6.

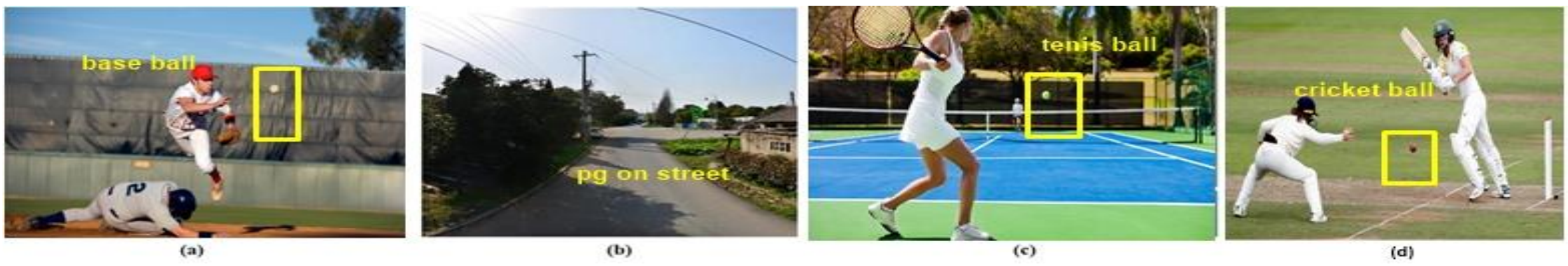

Fig. 1. Some instances examples of tiny objects

Table 5: A taxonomy of tiny object detection methods we discussed

| Type | Sub category | Method | Ref# | Type | Sub category | Methods | Ref# |
|---|---|---|---|---|---|---|---|
| Multi-scale feature learning (see section 2.3.1) | Featurized image pyramids | SNIP | [172] | Training strategy (see section 2.3.3) | | | [172] |
| | | | | | | SNIP | |
| | | CasMaskGF | [179] | | | SNIPER | [180] |
| | Single feature map | Fast R-CNN | [146] | | | SAN | [176] |
| | | Faster R-CNN | [148] | Context-based detection (see section 2.3.4) | Local context | MRCNN | [147] |
| | | | | | | MPNet | [153] |
| | | SPPNet | [143] | | | GBDNet | [154] |
| | | R-FCN | [151] | | | ACCNN | [162] |
| | Pyramidal feature hierarchy | SSD | [158] | | | CoupleNet | [164] |
| | | MSCNN] | [152 | | | SCAN | [170] |
| | Integrated features | ION | [157] | | Global context | ION | [157] |
| | | HyperNet | [150] | | | R-FCN++ | [171] |
| | Feature pyramid network | FPN | [161] | | | DeepIDNet | [145] |
| | Variants of FPN (including feature fusion and feature pyramid generation, multi-scaled fusion module, etc.) | RefineDet | [169] | | | SegDeepM | [149] |
| | | M2Det | [181] | | | CPF | [155] |
| | | DFPN | [178] | | Context interactive | SMN | [160] |
| | | FFSSD | [166] | | | ORN | [173] |
| | | FSSD | [168] | | | Context-SVM | [144] |
| | | DSSD | [159] | | | | |
| | | MDSSD | [182] | | | SIN | [174] |
| Data augmentation (see section 2.3.2) | | Augmentation | [183] | GAN-based detection (see section 2.3.5) | | Perceptual-GAN | [163] |
| | | | | | | MTGAN | [177] |

Table 4. Historical Summary of recent and old tiny object detection methods

| Year | Paper short briefed tittle | Method used | Methods (order from left to right) |
|---|---|---|---|
| 2014–2016 | Rich feature hierarchies analysis for accurate object detection and semantic segmentation | R-CNN [142] | |
| | Spatial pyramid pooling operations in deep convolutional networks for visual recognition | SPPNet [143] | R-CNN➔ SPPNet ➔ Context-SVM ➔ DeepIDNe➔ |
| | Contextualizing object detection and classification DeepID-Net: deformable deep convolutional neural networks for object detection | Context-SVM [144] DeepIDNet [145] | Fast R-CNN➔ MRCNN ➔ Faster R-CNN ➔ SegDeepM |
| | Fast R-CNN | Fast R-CNN [146] | |
| | Object detection by process of multi-region & semantic segmentation-aware CNN model | MRCNN [147] | |
| | Faster R-CNN: for real-time object detection with region proposal networks and some other techniques | Faster R-CNN[148] | |
| | segDeepM: manipulating segmentation and context in deep neural networks for object detection purpose, | SegDeepM [149] | |
| 2016–2017 | Hypernet: for correct region proposal generation and joint object detection | HyperNet [150] | HyperNet ➔ R-FCN ➔ |
| | R-FCN: object detection with region-based fully convolutional networks | R-FCN [151] | MSCNN ➔ MPNet [➔ |
| | An integrated multi-scale deep convolutional neural network for fast object detection | MSCNN [152] | GBDNet ➔ CPF ➔ |
| | A MultiPath network for object detection | MPNet [153] | R-CNN-SOD ➔ ION ➔ SSD |
| | Gated bi-directional CNN for object detection | GBDNet [154] | |
| | Contextual priming and feedback for faster R-CNN | CPF [155] | |
| | R-CNN for tiny object detection | R-CNN-SOD [156] | |
| | Inside-outside net: detecting objects in context with skip pooling and recurrent neural networks | ION [157] | |
| | SSD: single shot multibox detector | SSD [158] | |
| 2017–2018 | DSSD: Deconvolutional Single Shot Detector | DSSD [159] | |
| | Spatial memory for context reasoning in object detection | SMN [160] | DSSD ➔ SMN ➔ |
| | Feature pyramid networks for object detection | FPN [161] | FPN➔ ACCNN ➔ |
| | Attentive contexts for object detection | ACCNN [162] | Perceptual-GAN➔ CoupleNet➔ |
| | Perceptual generative adversarial net-works for tiny object detection | Perceptual-GAN[163] | SOD-Faster-R-CNN ➔ FFSSD ➔ |
| | CoupleNet: global coupling structure with local parts for object detection | CoupleNet [164] | ISOD➔ FSSD |
| | A closer look: small object detection in faster R-CNN | SOD-Faster-R-CNN [165] | |
| | Feature-Fused SSD: Fast Detection for Small Objects, CoRR abs/1709.05054 2017. | FFSSD [166] | |
| | Improving small object detection, | ISOD [167] | |
| | FSSD: Feature Fusion Single Shot Multibox Detector | FSSD [168] | |
| 2018–2019 | Single-shot refinement neural network for object detection | RefineDet [169] | |
| | SCAN: semantic context aware network for accurate small object detection | SCAN [170] | RefineDet ➔ SCAN➔ |
| | R-FCN++: towards accurate region-based fully convolutional networks for object detection | R-FCN++ [171] | R-FCN++ ➔ SNIP ➔ |
| | A vital analysis of scale invariance in object detection SNIP | SNIP [172] | ORN ➔ SIN ➔ |
| | Relation networks for object detection, | ORN [173] | Deconv-R-CNN ➔ SAN ➔ |
| | Structure inference net: object detection by using scene-level context and instance-level relationships, | SIN [174] | MTGAN ➔ DFPN ➔ |
| | Deconv R-CNN for small object detection on remote sensing images, | Deconv-R-CNN [175] | CasMaskGF ➔ SNIPER |
| | SAN: learning relationship between convolutional features for multi-scale object detection, | SAN [176] | |
| | SOD-MTGAN: small object detection via multi-task generative adversarial network | MTGAN [177] | |
| | Small object detection using deep feature pyramid networks, | DFPN [178] | |
| | Cascade mask generation framework for fast small object detection | CasMaskGF [179] | |
| | SNIPER: efficient multi-scale training, | SNIPER [180] | |
| | M2Det: a single-shot object detector based on a multi-level feature pyramid network, | M2DNet [181] | M2DNet ➔ |
| | MDSSD: a multi-scale deconvolutional single shot detector for small objects, | MDSSD [182] | MDSSD ➔ |
| | Augmentation for Small Object Detection | Augmentation [183] | Augmentation➔ |
| 2019-2020 | An improved faster R-CNN for small object detection, | Improved-Faster-R-CNN-SOD [184] | Improved-Faster-R-CNN-SOD |

### *2.3.1. Multi-scale feature learning*

Several scaling methods are used for feature learning. There are seven basic problem paradigms: featured image, single feature map, integrated features, hierarchical features, feature network, feature-based modules, feature fusion, multi-scale modules. In Fig. 2, we presented these complex methods and components by replacing them with CNN's resulting in vastly improved object detection performance. The single scale detector was used by Liu et al. to find an accurate scale for all of the images. The team found that training a single-scale detector was more difficult than training scale-dependent detectors with image pyramids to recognize tiny objects. Thus, they used a new scaling method for image pyramids (SNIP). We trained multiple scale-dependent detectors. Each of them is responsible for detecting specific items. The cascade mask framework incorporates multi-scale inputs for rapidly finding tiny objects.
However, computation-intensive algorithms require rapidly increasing memory and inference time. Regular computing detectors, such as Fast R-CNN [146], SPNet [148], and R-FCN [143], rely on full feature maps computed by CNNs, with different aspect ratios and scales (see Fig. 2(b) for details). However, the topmost objects have a conflict with each other due to their fixed receptive field.

Objects segmentation at different hierarchical levels for information capturing is considered the backbone of OD methods. Spatial-rich features can enhance the detectability of tiny objects in shallow layers. Only dense parts with semantic information capture have core resolutions but large target objects. In-depth CNN's produce more feature map resolutions in the in-line and create significant layers at different semantic depths (Fig. 2(c)). Liu et al. [91] discovered layers with different scales and aspect ratios, and due to his discovery, they were predicted from multiple levels of objects after that. The shallower layers used features for smaller objects, while the deeper ones served a different purpose. Developed by Cai et al. [152], MCCANN used multiple-resolution deconvolution layers and then produced their results using their refined feature maps see Fig. 2(d). S.Bell et al.[157] presented ION, which shortened multi-scale region featured through ROI pooling and merged them to produce the final region predictions. They created a hierarchical network with the name HyperNet. They made high-resolution hyper features by utilizing low and medium-resolution layers. The Feature-pooling map is better for localization and classification because it has elements from multiple levels of the image. The combined feature map increases memory but decreases performance; for more understanding, please see Table 5.
Cao et al. [166] proposed a fusion method known as feature-using SSD (FFSSD) for accuracy improvement of tiny object detection. In this process, the author used the concatenation module and sum module to fuse SSD's pyramidal detection method features. Li and Zhou presented a feature fusion single-box design (FSSD) hybrid device [168]. Components concatenating from multiple levels by down-sampling and feeding to multi-box detectors can produce pyramid feature sets and single-shot deconvolution (DSSD). Using skip connections and deconvolution layers, the performance of their dense feature maps DFS Leverages can be improved. The deconvolution layer and FPN for tiny objects (DFS and FPN in order of losses) also show better results. The systems that use fusion features can have an extra computational cost for every prediction layer. Unlike these architectures, X. et al.[154] developed the multi-scale deconvolution single-shot detector (MSSD), which started with SSD. The scale uses a multi-Scale deconvolution algorithm to achieve the low-level semantic features shown in Fig. 2(e,f,g). Moreover, they added unnecessary conv3 three layers in the middle of the neural network used to increase the performance of tiny object detection. Based on these concepts, we can use multi-scale learning to detect tiny objects.

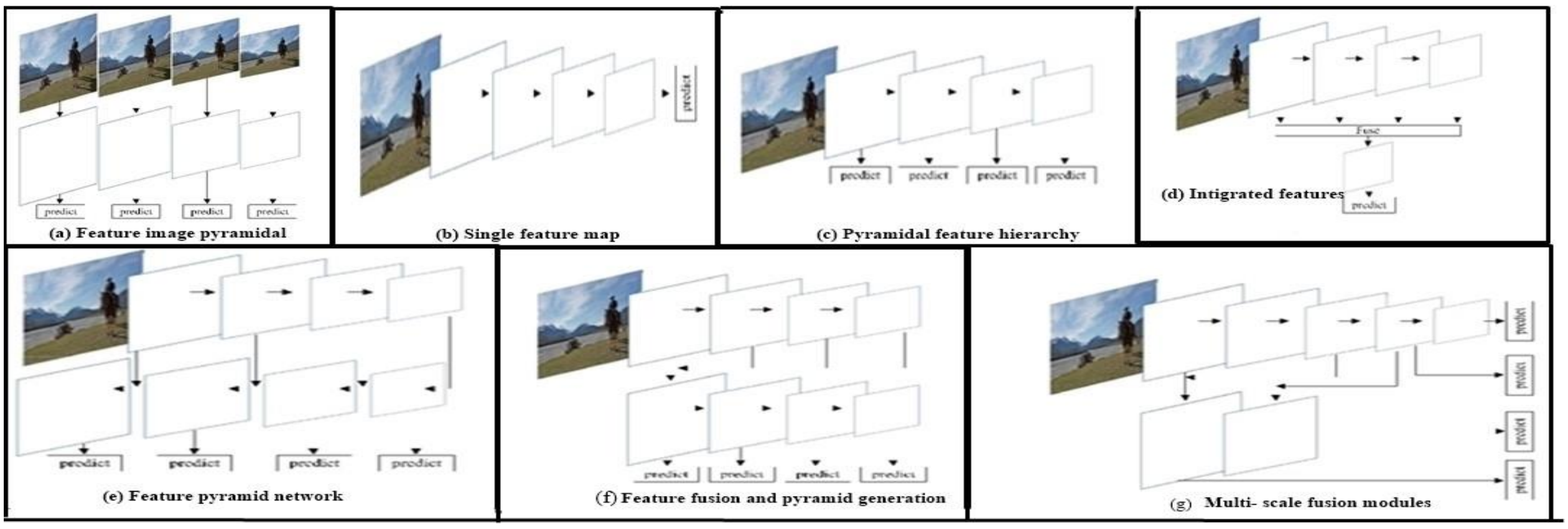

Fig. 2. Seven postulates for multi-scale feature learning.

**Note about Fig 2**. (a)It takes time to build a feature pyramid using an image pyramid because features are computed independently on each image scale. (b) Detection systems like Faster RCNN use only single-scale features (the last Conv layer) for faster detection. (c) An alternative to the featured image pyramid is to predict each of the pyramidal feature hierarchies from a CNN. (d) It refers to indicate on a single feature map generated from multiple features. (e) Feature pyramid network (FPN) integrates the structure of (b), (c), and (d). (f) Features from different layers with different scales are concatenated and used to produce a pyramid of features later. (g) Multi-scale fusion module with skip connections.

### *2.3.2. Data augmentation*

Data augmentation is the process of changing an image through transformations such as flipping, cropping, rotating, scaling, and so on. Data augmentation aims to generate more samples of class without changing the underlying category. Augmenting data can be used in training, testing, or both. In general, we can say, by using a large amount of data, the performance of deep learning can be improved. Similarly, increasing the types and numbers of tiny object samples in the dataset can improve tiny object detection performance. Kisantal et al. [183] investigated the problem of tiny object detection and thoroughly examined the state-of-the-art model called Mask-RCNN on the MS-COCO dataset. They showed that a lack of tiny objects in a training set is one of the main contributing factors to poor performance for tiny object detection. The reason behind it was only a few images contain tiny objects, and even within each image contain many tiny objects, and tiny objects do not appear frequently. Existing object detectors, in particular, require the presence of enough objects for predicted anchors to match during training. Kisantal et al. [183] proposed two approaches for augmenting the original MS-COCO dataset to address this issue. During training, they imagined that oversampled images contain tiny objects can quickly improve the detection performance of tiny objects. Secondly, they present an augmentation method that involves extracting tiny objects from images using a segmented mask and then copy-pasting tiny objects. In general, data augmentation improves the detection performance but computation complexity increase during the training and testing process and caused harmful effects on real-world applications. (see Table 5 for more info).

### 2.3.3. Training strategy

In general, relatively tiny- and large-sized objects are hard to detect on different scales. Singh and Davis[] created a gradient learning strategy named 'normalization for objects' to solve this problem. The purpose of this strategy was front-propagation, the generalization terms of image representations on relatively large and tiny scales (or comprehensive) instances. Specifically, the ground truth box is used to train every classifier, not just those associated with projects. They only select the lowest resolution ground truth boxes and only accept proposals that fall within a defined range during training. Thus, all ground truth anchors are also being used for RPN training. These anchors with ground truth overlap> 0.3 must be disregarded. A mechanism is to randomly generate alternative proposals for each resolution and categorize them independently in the testing stage. Meanwhile, they only search on the results, which fall within a range of values at each level of quality. The final results are obtained after bounding regression and classification techniques. They later developed SNIP [180], another multi-scale training strategy. In this case, the software only took care of regions around ground truth instances instead of the entire image pyramid. It used a multi-scale image pyramid for accelerating low-resolution training. In addition, they designed a channel-aware network known as the SAN and constructed a novel conceptually new network representation that focuses on the relationships. Various CNN features can be mapped to a scale-invariance subspace that can increase CNN detectors' robustness. During feature normalization, the network must first strip away any other spatial information, and after feature extraction, the network and feature extraction must be trained concurrently. These training strategies may help to find tiny objects, but it's mostly a challenge for detectors.

### 2.3.4. Context-based detection

Context plays a vital role in object detection. In general, visual objects commonly coexist with other visible entities, for example, many birds fly in the air same time. Contextual information is implicitly learned from a hierarchy of multiple levels of features. However, we still have good reason to look at contextual information explicitly. Consequently, visual context facilitates object recognition, especially for discovering and identifying tiny objects with few cues.

Furthermore, the R-CNN [156] was augmented with the proposal network to improve object detection of tiny objects. What you should do to learn from a single scene in a global context?? considered a big question in this domain. For example, suppose contextual information is included. In that case, finding a baseball becomes more accessible (e.g., baseball field, bat, and players)—several deep learning-based detectors for object recognition help to capture this global context. Recently, context-dependent object detectors can be considered our main goal here. The basic idea of this concept can be divided into two kinds: exploring relationships between individual items and modeling dependencies between items and contexts. A node in a relationship named SIN was thought of as a graph edge. By using contextual information in SIN detection ability was improved. Song et al. [144] proposed a contextualized learning method that features an adaptive and customizable model complexity into meaningful information.Meanwhile, they implemented a context-driven classification and detection algorithm that was more advance in contextual information extraction. A spatial memory network was created by using object instances and obtained spatial memory knowledge at the SMN level. Another auther discovered a lightweight object network (LWO), which stated the correlation between different parts of context; for more understanding, please see Table 5.

### 2.3.5. GAN-based detection

In few years, the generative adversarial networks (GAN) proposed by Goodfellow et al. [185] have gotten a lot of attention. The two-person zero-sum game in game theory is the structural inspiration for GAN. A standard GAN consists of a generator network and a discriminator network competing in a minimax optimization system. The generator learns to capture the possible distribution of valid data samples and produces new data samples. At the same time, the discriminator attempts to distinguish between the right data distribution instances and those produced by the generator. To increase our knowledge, Li et al. [163] proposed a novel perceptual GAN model that was the first to apply GAN to an object detection task. Its purpose was to improve tiny object detection efficiency by generating super-resolved representations for tiny objects and reducing the representation gap between tiny and large objects.Meanwhile, the discriminator on the generator imposes a perceptual constraint on the produced representations: that must be helpful to detect tiny objects. Bai et al. [177] proposed MTGAN, a multi-task generative adversarial network, to solve tiny object detection. The MTGAN's generator is a high-resolution network that up samples tiny blurry images into fine-scale consistent photos. The discriminator in the MTGAN is a multi-task network, unlike the generator. Each super-resolved image patch in the discriminator is represented by a natural or fake score, object category scores, and regression offsets. Furthermore, during testing, the discriminator bounding box regression and classification losses are back-propagated to the generator, allowing the generator to obtain more data for more accurate detection. Extensive tests show that the above two GAN-based detection methods outperform state-of-the-art algorithms in detecting tiny objects like traffic signs; for more conceptual ideas, please see Table 4,5.

## 2.4. General state-of-the-art of famous face detection techniques

Face detection gained a lot of attention and achieved a lot of success. Faces have distinct facial features, such as the nose, eyes, mouth, and relative positions, combining to form a face detection distinct from generic object detection. However, it is not easy when face sizes are tiny, for example, less than 16x16 pixels, and features are indistinguishable. Many techniques for detecting small faces have been developed. Table 6 displays the most recent state-of-the-art and the result of significant algorithms.

Table 6: Face Detection strategies with one/two stage models, mean-average-precision(mAP)

| | | | | Wider Face validation Set | | |
|---|---|---|---|---|---|---|
| **Network** | **One-Two stage** | **Backbone** | **Strategy** | Easy (mAP) | **Medium (mAP)** | **Hard (mAP)** |
| Tiny Face | One-stage | ResNet101 | Context reasoning | 0.919 | 0.908 | 0.823 |
| SSH | One-stage | VGG16 | Feature fusion/context reasoning | 0.931 | 0.921 | 0.845 |
| SRN | Two-stage | ResNet50 | Anchor matching /balance classes | 0.957 | 0.946 | 0.884 |
| S3FD | One-stage | VGG16 | Anchor matching /balance classes | 0.937 | 0.924 | 0.852 |
| EXTD | One-stage | Inverted Residual | Feature fusion | 0.912 | 0.903 | 0.85 |
| DF2S2 | One-stage | ResNet50 | Feature fusion/context reasoning | 0.969 | 0.959 | 0.912 |
| PyramidBox | One-stage | FPN | Feature fusion/context reasoning | 0.961 | 0.95 | 0.889 |
| PyramidBox++ | One-stage | FPN | Context reasoning/balance classes | 0.965 | 0.959 | 0.912 |
| FA-RPN | One-stage | RPN | Anchor matching /balance classes | 0.95 | 0.942 | 0.894 |
| Face-MegNet | Two-stage | VGG16 | Context reasoning/feature fusion | 0.92 | 0.913 | 0.85 |
| SFA | One-stage | VGG16 | Anchor matching/feature fusion | 0.949 | 0.936 | 0.866 |
| Face-RCNN | Two-stage | VGG19 | Anchor matching /balance classes | 0.938 | 0.922 | 0.829 |
| RetinaFace | One-stage | – | Context reasoning | 0.969 | 0.961 | 0.92 |
| RAP | One-stage | Dilated convolutional | Anchor matching | 0.949 | 0.935 | 0.865 |

### 2.5. Tiny object detection challenges and deep domain adaptation

Different autoencoders can accomplish the reconstruction of target data from source data. In the case of tiny object detection, it can be considered a more challenging task. In many cases, object mismatching and shuffling with background chances increases as compared to general object detection. Identification of object domain and data reconstruction in nonclear objects is also considered a big challenge. This survey paper reviewed both subjective domains (tiny objects and domain adaptation) and highlighted different techniques that can help to solve these challenges. We examined different object detection models and mapped on tiny object detection as well as domain adaptation techniques. After implementing OD models on different techniques, we obtained results with different datasets, as shown in Table 3 and Section 5.

Autoencoders are also considered a form of neural network that helps reconstruct input/output from hidden layers. In general, we can present it:$||\mathbf{k\psi(\psi(xM)M^{-1})-xk}||$, where M is matrix projection and ψ is nonlinear activation function[517]. There are several approaches to avoid with simplistic solutions for the challenge. One formula extracts important data by pressing the input through a bottleneck (contractive auto-encoders). At the same time, another also introduces fake noise to the input that the network needs to remove (denoise autoencoders)[518]. Deep auto-encoders can stack several nonlinear layers to provide flexible transformations [519],[520]. However, stacking numerous layers can raise computational costs, but the denoising cost of encoders can be reduced by marginalizing noise [521]. Autoencoders are not the only neural networks used for domain adaptation and object reconstruction, and different DDA techniques help for more such problems. The Domain-adversarial neural network (DANN), one of the best-known adaptive networks, seeks to establish a representation that allows the domains to classify source samples for small/large objects [119]. This function is done with the help of two-loss layers: one loss layer classifies samples by label, while other loss layers classify data by domain. During optimization, DANN minimizes the loss of label classification and maximizes the loss of domain classification. Adverse domain networks rely mostly on generalization error: if the domain discrepancy is small, the source classifier's target error will be small. But in the case of tiny objects, domain discrepancy increases sometimes. That's why target data error chances for the source classifier also increases. Maximizing the sample classification error of distinct domains is similar to the proxy distance being minimized as eq(0):

$$D_A[x, z] = 2\,(1 - 2\,\hat{e}\,(x, z)) \quad \text{eq(0)}$$

Where ê (x, z) is considered an error of cross-validation for the classifier, it is trained to discriminate source sample x from target samples z [522]. The A-distance proxy is obtained from the total distance of variation [522]. Different distances are used to classify domains, particularly concerning moment matching [523] or the use of Wasserstein distance [524]. These could be costly computationally, and less expensive methods are being developed, such as the central moment discrepancies or the pairs of hypotheses[525][526].

A shortcoming of the domain-adversarial networks is that matched data distributions do not imply matching class-conditional distributions, as mentioned earlier in the discussion. In the case of tiny objects, the mismatching of class conditional distribution increases. The reason behind this is considered finding a lot of similar unclear objects at the same time. In addition, the two-loss layers generate gradients that are often distinct. This situation makes it more difficult to train DANNs than normal deep neural networks. However, some new work has also previously been examined to stabilize the learning process. The DIRT-T technique speeds up gradient descent by utilizing the natural gradient to prevent high-density data areas from crossing their decision boundary[527]. Substituting domain confusion that maximizes the objective with its dual formulation can solve this problem[528].

Several network architectures such as residual layers[529] explore the idea of maximizing domain confusion while minimizing misclassification even object is smaller or larger [530][531]. Furthermore, it helps in tying weights to different levels[532], kernel embedding to higher layers [532], aligning moments [533], and modelling domain-specific subspaces [533]. It is also used in several issue domains, including speech recognition[534], medical image segmentation[536], and the cross-lingual communication of knowledge[535]. We refer to further detailed study about domain adaptation and tiny object detection [537-539].

#### *2.5.1. Deep domain adaptation application: Tiny Object detection*

Convolutional neural networks with regions are driving recent advancements in object detection (R-CNNs [541], fast RCNNs[540] and faster R-CNNs [542]). They are made up of a window selection mechanism and classifiers labelled bounding boxes that have been pre-trained using CNN features. The classifier determines if an object is present in a region produced by sliding windows at test time. Although the R-CNN method is effective, each detection category requires a considerable amount of bounding box labelled data to train. In tiny objects detection, this problem exists more due to unclear objects. Labelling and sliding windows production for tiny objects also complex for the source as well target domain. Deep DA approaches can be employed in classifiers to adapt to the target domain to tackle the lack of labelled data because the window selection mechanism is domain-independent. Weakly labelled data (such as image-level class labels) are directly useful for the detector because R-CNN's train classifiers on regions similar to classification. The detector usually learns with a small amount of bounding box labelled data and a large amount of weakly labelled data. The large-scale detection via adaptation (LSDA) [544] technique trains a classification layer for the target domain. It then updates the target classification parameters directly using a pre-trained source model and output layer adaptation techniques. Rochan et al. [543] employed word vectors to build semantic relatedness between weakly labelled source and target items, then transferred bounding box labelled information from source to target objects depending on their relatedness. Tang et al. [546] extended [544] and [543] by transferring visual (based on the LSDA model) and semantic (based on work vectors) similarities for training an object detector on weakly labelled categories. [547] used adversarial training to reduce the domain disparity by incorporating both an image-level and an instance-level adaption component into faster R-CNN. [545] fine-tuned the pre-trained model with domain-transfer samples and pseudo-labelling samples using bounding box labelled data in a source domain and weakly labelled data in a target domain.

## 3 OBJECT DETECTION TRENDING ARCHITECTURES

Several convolutional methods have been discovered that offered outstanding contributions in different detection needs, following procedures performing a significant role in detecting state-of-the-art these days.

Famous object detection methods that are getting recent research trends are (Convolutions, Convolutional Neural Networks, Pooling Operations, Normalization, Skip Connection Blocks, Feature Extractors, Skip Connections, Image Model Blocks, Feedforward Networks, Regularization, Feature Pyramid Blocks, Feature Extractors, Initialization, Activation Functions, Instance Segmentation Models, Learning Rate Schedules, RoI Feature Extractors, Region Proposal, Stochastic Optimization, Output Functions, Regularization)

Due to keeping in mind the length of the paper, we will only elaborate few of them for the object detection survey.

### 3.1. Convolution methods and their types for object detection: Image processing

It is a type of operation that support learning patterns from displayed image. After learning, a kernel slides over an image and performs element-wise multiplications within input in this method. It allows sharing parameters and translation of invariances. This method performs best supports in image feature extraction. Another way to define convolutions is a type of matrix operation performed with the consistency of a kernel and tiny weight matrix, element-wise multiplication performed with kernel help using input data operations summary of results prepared into an input Fig 3. Best operations performed by Convolution can be expressed intuitively. It allows sharing weight, reducing adequate parameter numbers, and translating images (it detects the same type of features in different input data) [186]. This method for image feature extraction started in 1998, and till now, almost 6751 papers published on this topic. Following are different types of convolutions

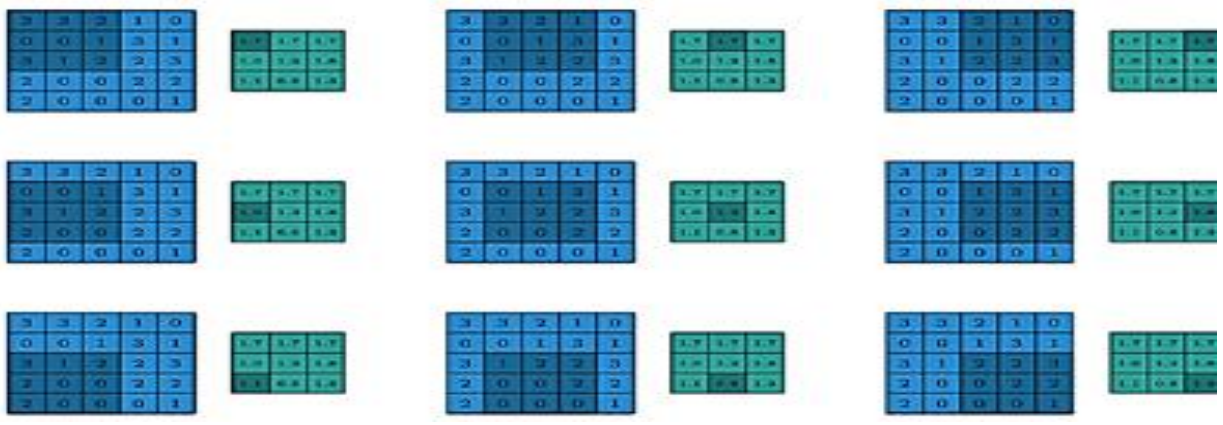

Fig 3. Convolution arithmetic for deep learning, matrix operation [186]

#### *3.1.1. Types of convolution methods.*

Following are different trending types of Convolution, and most of them are specially considered in image processing techniques.

***a. 1x1 Convolution(1x1C):*** This method started in 2013; this type of Convolution was used for dimensions reduction with the embedding process low-dimensional properties. This mechanism helps the nonlinear process after Convolution. It supports squeezing input pixels into output pixels with all of the helping channels. In general, it is a machine learning process used for particular pixel location identification shown in Fig 4.a & [187,188].

***b. 3D Convolution(3DC):*** This method started in 2015; it is a method in which kernel sliding was performed using the 3D convolutional and dimensional process opposing two-dimensional and 2D convolutions. For example, in the medical imaging process, model construction is performed using 3-dimensional image slices. This model is suitable more for video-based data as in this data, more dimensional Tempore present [186]—comparison of 2D and 3D convolution shown in Fig 4.b.

***c. Active Convolution (AC):*** This method started in 2017. We found two popular papers published using this method. In this convolution type, fields do not have definite shapes; this method can draw various related fields for convolutions. Backpropagation support this process for shapes learning during training. It is the general form of Convolution, not only convolutional Convolution as well as fractional pixels coordinates. It provides greater freedom to convolutional neural network structure management, as we can change the shapes of convolutions freely. Another thing, shapes of convolutions can be altered by training instead of by hand [112] [113]. Comparison of convolutional convolution unit with the ACU. Fig 4 c.(a)This is convolutional convolution within two input neurons and four output neurons. Fig 4 c(b)Similarly convolution unit, the synapses of the ACU can be connected at inter-neuron positions and are moveable.

***d. Attention-augmented Convolution (AAC):*** This method started in 2019, and we go through 1 paper that showed performance by using this object detection method. AAC is a type of convolution with a two-dimensional relative self-attention mechanism. It can replace convolutions with a stand-alone computational primeval that can be used for image classification. Working in this method is performed by concatenation convolution and attentional map of features. AAC is equal to translation and also always ready to operate in different multi-dimensional environments shown in Fig 4.d & [114].

$$\text{AAConv(X)} = \text{Concat[Conv(X),MHA(X)]} \ldots\ldots\ldots\ldots Eq1$$

This method works with the help of convolutional and attentional feature map concatenation regional convolution operator help to see this with kernel size input filter and outputs filters. It can be written as X originates from an input tensor of shape$(H, W, F_{in})$. This is flattened to become$X \in \mathbb{R}^{HW \times F_{in}}$ .Which is passed into a multi-head attention module, as well as a convolution (see above Eq1). So, it can be said AAC is equal to translation and operate readily on different spatial dimensions.

**e. *Conditional Convolution (CondConv):*** This method started in 2019, and we found two best papers that discussed this OT method. In this type of convolution, specialized kernel work for each working part. Cond Conv layer as a linear combination of n experts $(\alpha_1 W_1 + \cdots + \alpha_n W_n) * x$, here α1,…,αn are function that learns from gradient descent of input. The model's efficiency can be improved by increasing the number of layers in Cond Conv, the same as efficiency can be improved by increasing the number of developers in any system. Instead of increasing convolutional kernel size, that model increases efficiency by applying Cond Conv on different input positions shown in Fig 4.e & [115] [116]. Fig 4. (e)a. Conditional convolution layered architecture with n=3 kernels vs. Fig 4. (e)b. a mixture of several expert approaches by doing parameters of the convolutional kernel according to condition on input. CondConv is equal to the mix of experts approaches mathematically but need only one convolution.

***f. Convolution:*** This is the primary method, but we also mentioned it here according to the symmetrical presentation. Convolution started in 1980. Several researchers offered a lot of knowledge by using this method. In this type, matrix operations are performed with the consistency of a kernel and little matrix of weights that slide within input data, perform element-wise multiplication, and summarise all results. It serves an essential role in weight sharing, parameter reduction, and image translation [187]Fig 3.

patch input, A group of kernel offsets Δk generated by the deformable kernel by patch feature input with the help of light-weight generator G that is

***g. Coordinate Convolution (*CoordConv*):*** This method started in 2018, and we go through 6 published papers that used this method for object detection and image processing. CoordConv layer is the most straightforward extension of layer convolutional. Its functional signature is the same as layer convolutional, but the task is fulfilled by concatenating the first upcoming new extra channels; there is a hard-coded coordination system in these channels. In the basic version of this system, one channel is for i-coordinate and one channel for j-coordinate. CoordCov has properties of efficiently computing on few parameters from convolutions, but it allows the network to keep translation invariance for learned tasks. It is best practice to coordinate transformation task at a time when regular transformation fails [205][206]Fig 4 (g). A comparison of 2D convolutional and CoordConv layers can be seen. On the left side, there represented a slandered form of convolutional layer maps with the representation block h* w*c to describe a shape of h'*w'* c'. A CoordCon layer has the same functional signature on the right side, but its complete mapping by first concatenating extra channels for the incoming representation. In these channels, there present hard-coded coordinates; in the most basic version for them, there is current i-coordinate one channel and for j-coordinate one medium that can be seen in Fig 4 (g). Other derived coordinate may be input same as well, like radius coordinate used in ImageNet experiments.

***h. Deformable Convolution (DC):*** This method started in 2017, and we found almost 47 papers that used DC. In this method, convolution added by 2D offsets in the regular grid sampling location in standard convolution. By this process, deformation of sampling grip enabled [207]. From the preceding feature maps, the offsets get learned with the help of extra convolutional layers. This method deformation is conditioned on the input features in an adaptive, dense, and local manner [208] Fig 4(h).

***i. Deformable Kernel (DK):*** This method started in 2019. We have seen the two most reputed papers that used this method for their results. This type of convolutional operator is used for deformation modeling. DKs learn free-form offsets on kernel coordinates to deform original kernel space towards specific data modality instead of data reposting. By this method, directly effective receptive filed (ERF) can be adopted while have to leave untouched receptive filed. In rigid kernels, these can be used as a drop-in replacement. In Fig 4(i), it is represented, for example, of a*3*3 convolution of the rigid kernel. Original given kernel weight W and offset group, Deformation kernel is a sample of a new set of kernels W^' with the help of bilinear sampler. Finally, Deformation kernels convolute input feature maps and kernels sampled to complete all computation processing [209].

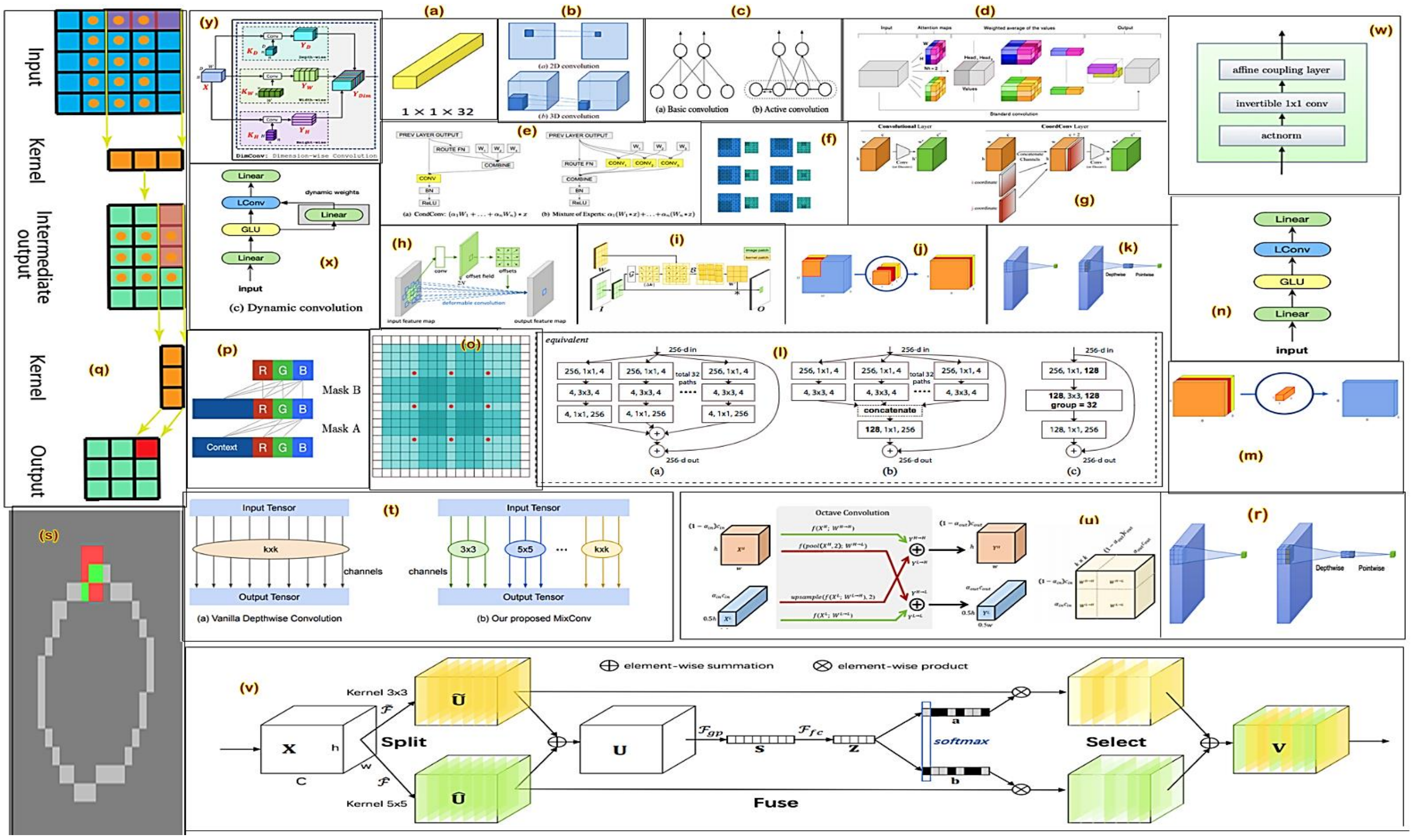

**Fig 4.** Common types of Convolution methods, (a-u, different diagrams about these methods presented at exact place to have overview about methods and cited in above-explained contents)

### *3.1.2. Summary of other popular types of Convolution methods*

Now we will represent generally with the help of method name, method started year (msY) almost published papers using method(appM) and with the help of most recent published papers by using this method. (Depth wise Convolution msY (2016) appM (286) [210] Fig 4. (j)). (Depth wise Dilated Separable Convolution [189] msY (2018) appM (2) Fig 4.(k) )(Depth wise Separable Convolution [190] msY (2017) appM (251) Fig 4. (r)) (Dilated Convolution [191] msY (2015) appM (191) Fig 4. (o))(Dim Convolution [192] msY (2019) appM (1) Fig 4. (y))(Dynamic Convolution [193] msY (2019) appM (3) Fig 4.(x)) (Grouped Convolution [194]msY (2012) appM (422) Fig 4. (l)) (Groupwise Point Convolution [195] msY (2018) appM (21) Fig 4.(m))(Invertible 1x1 Convolution [196] msY (2018) appM (25) Fig 4. (w))(Light Convolution [197] msY (2019) appM (1) Fig 4. (n))(Masked Convolution [198] msY (2016) appM (9) Fig 4.(p)) (Mix Convolution [199]msY (2019) appM (3) Fig 4.(t))(Octave Convolution [200] msY (2019) appM (5) Fig 4. (u)) (Pointwise Convolution [201] msY (2016) appM (289)Fig 4. (m))(Selective Kernel Convolution [202] msY (2019) appM (4)Fig 4. (v))(Spatially Separable Convolution [203] msY (2000) appM (9) Fig 4. (q)) (Submanifold Convolution [204] msY (2017) appM (4)Fig 4. (s)).

## 3.2. Convolutional neural networks methods and types for Object detection

Convolutional Neural Network (ConvNet/CNN) is a deep learning algorithm that can take an image, assign importance (learnable weights and biases) to various aspects/objects in the image, and distinguish one from the other. When compared to different classification algorithms, the amount of pre-processing required by a ConvNet is significantly less. While filters in primitive methods are hand-engineered, ConvNets can learn these filters/characteristics with enough training. CNN working process can be seen in Fig 6.

With the Artificial Neural Network advancements, machine learning has taken a dramatic turn in recent periods (ANN). In everyday machine learning tasks, these biologically inspired computational models outperform previous forms of artificial intelligence. The Convolutional Neural Network is one of the most impressive types of ANN architecture (CNN). CNN's are primarily used to solve complex image-driven pattern recognition tasks, and their precise yet simple architecture provides a simplified way to get started with ANNs [211].

Deep Learning, also known as Deep Neural Network, refers to Artificial Neural Networks (ANN) with multiple layers. It has been regarded as one of the most potent tools in recent decades. It has gained wide acceptance in the literature due to its ability to handle massive amounts of data. Deeper hidden layers have recently begun to outperform traditional methods in various fields, most notably pattern recognition. The Convolutional Neural Network is a popular deep neural network (CNN). It gets its name from the linear mathematical operation between matrices known as convolution. CNN comprises several layers, including a convolutional layer, a non-linearity layer, a pooling layer, and a fully connected layer. Pooling and non-linearity layers do not have parameters, but convolutional and fully connected layers do. In machine learning problems, CNN performs a key role. Awe-inspiring applications deal with image data, such as the most significant image classification dataset (Image Net), computer vision, natural language processing (NLP), and get the final required results. This section will explain and define all elements and critical issues associated with CNN and how these elements work [212].

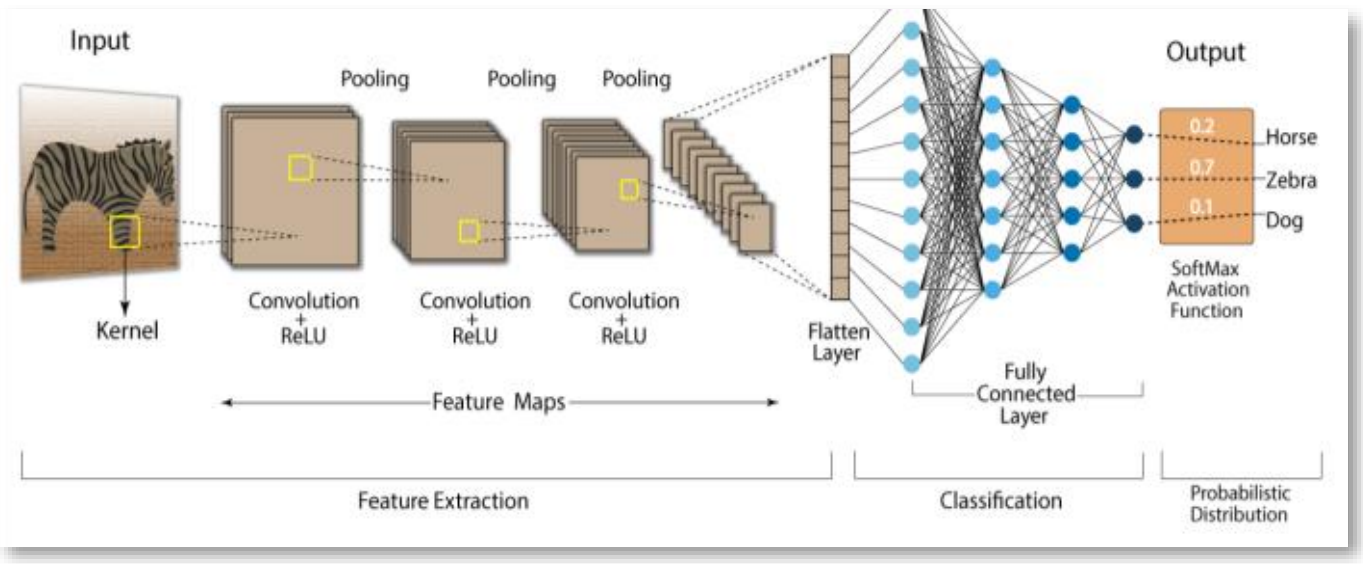

Fig 6. Convolution Neural Network (CNN) general example [516]

*(a) Residual Network (ResNet) [267]:* This method started in 2015, and almost 940 highly reputed research papers we found about it. ResNets type of method that learns function from layered reference input rather than unreferenced function learning. It fits directly on desired underlying mappings rather than hoping on few stacked layers; layers of residual nets include residual mapping. All layers form residual blocks on top of each other; for example, ResNet =50 consists of fifty layers using these blocks. Desired underlying mapping can be denoted as $\mathcal{H}(x)$; for different stacked nonlinear layers, mapping can be represented as $\mathcal{F}(x) := \mathcal{H}(x) - x$and recasting form as $\mathcal{F}(x) + x$. It is already proved these networks optimization is straightforward, and accuracy can be gained by a range of increased depth, shown in Fig 5(a).
*(b) AlexNet[213]:* This method started in 2012, and we checked almost 286 papers published in highly reputed journals by using this method. It is considered a classical type of CNN and consists of convolutions, max pooling, dense layers that work as basic building blocks, and grouped convolutions help model fitting across two GPUs [213]. Faults tolerance in Convolutional Neural Networks: FT-CNN used for finding spots during object recognition processing. Evaluation approaches were performed using different models like AlexNet, VGG-19, ResNet-18, and YOLOv2. This implementation handled soft errors with minimal 4% to 8% error-free and error injection situation [214], shown in Fig 5(b).
*(c) VGG[279]:* This started in 2014, and more than 244 papers we checked that published in high reputed journals, and it's a great challenge for researchers. With a minimal (3x3) convolution filter, considerable improvements can be attained by 16-19 weight layers [215]. This method mainly focuses on analytical experiments for networks depth increase ability. Two-compartment spiking neural network with superposition states encoding with inspiration quantum information theory proposed. This work helped inverted colour versions of character recognition and worked out from neuron information encoding from the brain. VGG, ANN, DenseNet and ResNet rolled with fully connected layers to accomplish this task. This proposed architecture named QS-SNN [216], shown in Fig 5(c).
*(d) DenseNet[288]:* This method started in 2016, and more than 200 papers we reviewed were published using this method. This type of CNN network manages dense connections between layers with the support of Dense blocks (DB). We can consider DB support directly in layers links by matching the size of features maps, same as feed forwarding [218]. Using EfficientNet architecture, some students performed plant pathology classification with the exploration standard benchmark, i.e. VGG16, ResNet101, DenseNet 161 and achieved a score of 0.945 [217], shown in Fig 5(d).
*(e)MobileNetV2[289]:* This method started in 2018, and more than 112 papers we reviewed were published in high reputed journals using this method. This type of Convolutional Neural Network performs a better role on mobile devices. Its structure is inverted residential connections type between bottleneck layers [219]. The intermediate layers use lightweight convolution features for the source of non-linearity. This method supported in COVID-19 for facial masks detection as a masked area detector from the collection of videos [220], shown in Fig 5(e).
*(f) GoogLeNet[293]:* In 2014, work started in this CNN method, and we go through more than 109 research papers that used this method and published them in highly reputed journals. This CNN based on Inception architecture. Several inception models utilized by this method help network to select many convolutional filter sizes in every block. Modules are stacked by inspection network with occasional max-pooling [221]. Several sensitivity measures are compared on VGG-16&GoogleNet trained on ImageNet/Places-365 datasets, which have been considered "Object detectors"[222]. In skin disease detection these popular deep learning networks i.e MobileNet, ResNet_152, GoogLeNet, DenseNet_121, and ResNet_101 with 89% results accuracy facilitation obtained [223] , shown in Fig 5(f).
*(g)ResNeXt[301]:* It started in 2016, and it repeats building blocks in the same topology with the aggregation of several transformations. As compared to ResNet, it controls the cardinality, i.e. dimensions (depth, width). It is represented as $\mathcal{F}(x) = \sum_{i=1}^{C} \quad \mathcal{T}_i(x)$, where $\mathcal{T}_i(x)$ is an arbitrary function. It is analogous to simple neuron, $\mathcal{T}_i$ Project $x$ into arbitrary dimensions [224]. Dynamically Throttleable Neural Networks (TNN) proposal validated on experimental validation using different Convolution Neural Networks (CNNs such as VGG, ResNet, ResNet, DenseNet) using CiFAR-10 and ImageNet dataset, the primary purpose of the application is object classification and recognition [225], shown in Fig 5(g).
*(h)Xception[305]:* Xception is a convolutional neural network architecture that relies solely on depthwise separable convolution layers [226]. Models of deep learning training on mobile phone-acquired frozen section images effectively can detect basal cell carcinoma. In the results, images were downscaled from a 4032 x 3024-pixel resolution to 576 x 432-pixel resolution. Deeplab V3 with Xception algorithms of semantic segmentation work as backbone was for model training. 2D black and white output can be predicted of the same dimension from images input inside the models, the area with basal cell carcinoma displayed as white color, in a black background [227], shown in Fig 5(h).
*(i)Darknet-53[310]:* This CNN was discovered in 2018, and we reviewed almost 58 papers. This work is the backbone for YOLOv3 object detection; improving its impotence are residual connections and more layers [228]—Darknet-53 is also useful for real-time instance segmentation and learning universal shape dictionary. Experimental results obtained on challenging COCO datasets and results using the single model on a single Titan Xp GPU achieve 35.8 AP and 27.8 AP at 65 fps with YOLOv4 as a base detector, 34.1 AP 28.6 AP at 12 fps with FCOS as base detector [229]. Rice disease can be detected via chatbot rice paddy service. The better results than previous contributions obtained, YOLOv3, trained by advanced training dataset. Performance measured on five target classes showed improved average mAP from 82.74% at an earlier paper to 89.10% [230], shown in Fig 5(i).
*(j) SqueezeNet[231]:* This convolutional neural network started in 2016; we studied it with the following outcomes.squeezeNet is a CNN that decreases the number of parameters by design strategies by using fire modules. As a result, parameters were squeezed by using 1x1 convolutions. Ongoing examination on profound neural organizations has zeroed in basically on improving precision. It is usually conceivable to recognize different DNN structures that accomplish that exactness level for a given precision level. With equal precision, more modest DNN structures offer in any event three preferences: (1) Smaller DNNs require less correspondence across workers during appropriate preparation. (2) Smaller DNNs require less data transfer to send out another model from the cloud to an independent vehicle. (3) Smaller DNNs are more doable to convey on FPGAs and other equipment with limited memory. To give these points of interest, we propose a small DNN design called SqueezeNet. SqueezeNet accomplishes AlexNet-level precision on ImageNet with 50x fewer boundaries. Moreover, with model pressure procedures, we can pack SqueezeNet to under 0.5MB (510x more modest than AlexNet)[231].SqueezNet also outperforms a significant role in the Grafted network for person re-identification. The rootstock can be based on the leading parts of ResNet-50 that provide a strength-based, while this is a newly designed module composed of the latter parts of SqueezeNet to compress the parameters [231] Fig 5(j).

### *3.2.1. Summary of popular CNN methods.*

*(k) Inception-v3[232]* Fig5(k) *(l) EfficientNet[233]* Fig5(l) *(m) LeNet[234]* Fig5(m) *(n) Darknet-19[235]* Fig5(n) *(o) MobileNetV1[238]* Fig5(o) *(p) WideResNet[236]* Fig5(p) *q) ShuffleNet[237]* Fig5(q).Some more types of CNN (r) SENet ,2017 Fig5(r) (s) MobileNetV3,2019 Fig5(s) (t) MnasNet,2018Fig5(t)(u)Inception-ResNet-v2,2016Fig5(u)(v)HRNet,2019.DPN,2017Fig5(v).ShuffleNetv2,2018.Inceptionv4,2016.InceptionV2,2015.OverFeat,2013.CheXNet,2017 PyramidNet,2016.CSPDarknet53,2020.AmoebaNet,2018.FractalNet,2016.RevNet,2017.FBNet,2018.SimpleNet,2016.SNet,2019.SpineNet,2019. ZFNet,2013.Single-path NAS,2019.DetNet,2018.MixNet,2019.

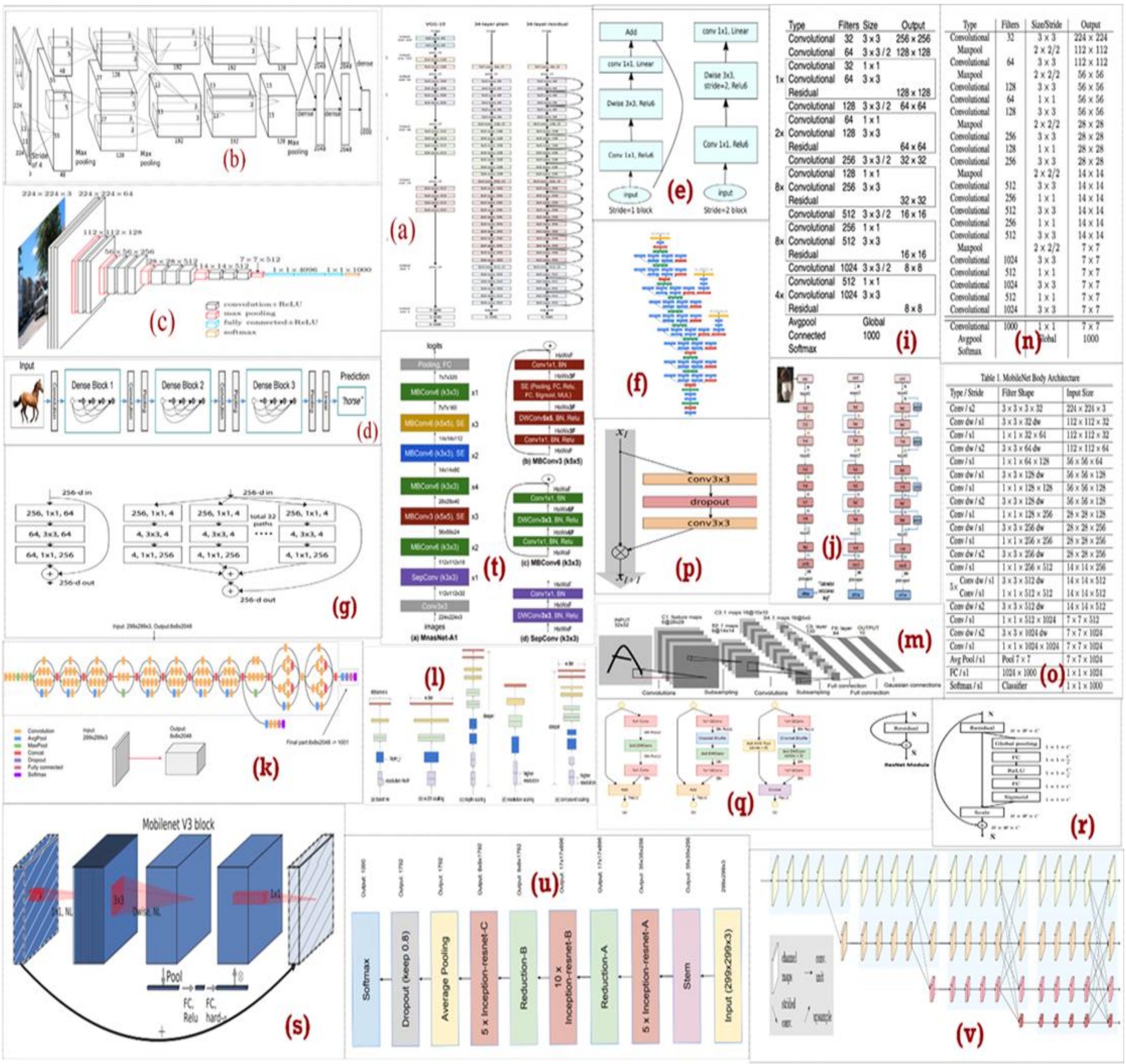

Fig 5. Common types of convolutional neural networks (CNN) methods suitable for object detection (a-v, different diagrams about these methods presented at the same place to have an overview of approaches and cited in the above-explained content).

## 3.3. Pooling operational methods and types for object detection

Pooling operations are needed to collectively pool different features; this method downsamples the feature map into smaller sizes. Pooling properties are induced in several other classifications, such as translation invariance of an image; it also supports information collection from different network parts for object detection (different scale pooling). Several types of pooling operations explained below concerning object detection [239]. In Table 7, we showed an overview of varying pooling operations.

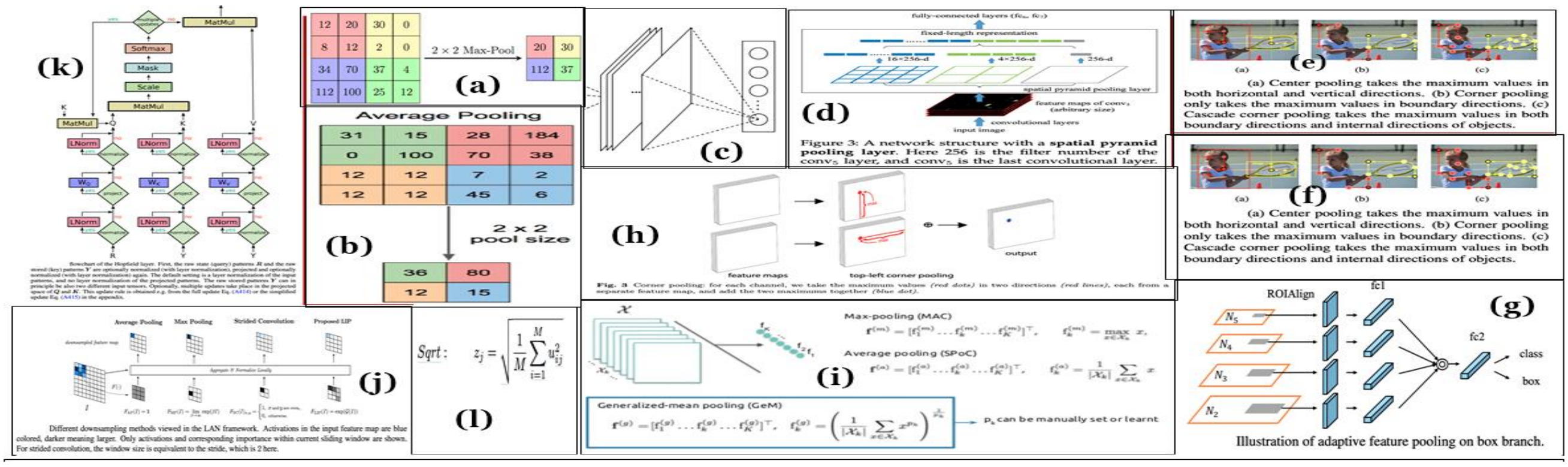

**Fig 7.** Pooling operation methods used for objects detection (a-l, different diagrams about these methods presented at exact place to have overview about methods and cited in above-explained contents)

Table 7. Different pooling operations for a role in object detection

| Method Name | Explanation Method Start IN(MS)Paper Reviewed (PR) | Ref# |
|---|---|---|
| ***a. Max Pooling*** ***MS(2000)*** ***PR(2405)*** | A pooling operation that determines the maximum value for patches of a feature map and uses it to construct a downsampled (pooled) feature map is max pooling. After a convolutional sheet, it's typically used. A small amount of translation invariance is added by this method, which means that small changes in the picture do not have a noticeable impact on the values of most pooled outputs Fig 7(a). | [240] [241] |
| ***b.Average Pooling*** ***MS(2000)*** ***PR(1618)*** | Average Pooling is a feature map downsampling (Pooling) process that helps in the calculation. For patches of a feature map, average values construct a downsampled (pooled) feature map. After a convolutional sheet, it's typically used. It also helps a small amount of translation invariance, which means that small changes in the picture have a substantial noticeable impact on the values of most pooled outputs. It extracts features more fluidly than Max Pooling, while Max Pooling extracts more prominent edges (Fig 7(b). | [242] [243] |
| ***c.Global Average Pooling*** ***MS(2013)*** ***PR(1313)*** | Global Average Pooling is a pooling process used in classical CNNs to replace linked layers entirely. In the last mlpconv sheet, the idea is to create one feature map for each category of the classification task by taking the average of each feature map used to feed the resulting vector directly into the softmax layer, rather than inserting completely connected layers on top of the feature maps Fig 7(c). | [244] [245] |
| ***d.Spatial Pyramid Pooling*** ***MS(2014)*** ***PR(71)*** | SPP (Spatial Pyramid Pooling) is a pooling layer that eliminates the network's fixed-size restriction, allowing a CNN to operate without a fixed-size input image. We apply an SPP layer, On top of the final convolutional layer. The SPP layer aggregates the features and produces fixed-length outputs fed into the fully connected layers (or other classifiers). To put it another way, we do some information aggregation at a higher level of the network hierarchy (between convolutional and fully connected layers) to avoid the need for cropping or warping at the start of Fig 7(d). | [246] [247] |
| ***e.Center Pooling*** ***MS(2019)*** ***PR(8)*** | Center Pooling is an object detection pooling strategy that seeks to capture more complex and identifiable visual patterns. Objects' geometric centres do not always express easily identifiable graphic designs (e.g., the human head contains solid visual patterns, but the centre key point is often in the middle of the human body) Fig 7(e). | [248] [249] |
| ***f.Cascade Corner Pooling*** ***MS(2019)*** ***PR(8)*** | Cascade Corner Pooling is an object detection pooling layer that extends the corner pooling process. Corners are often found outside of artefacts that lack local appearance characteristics. CornerNet employs corner pooling to solve this problem, which entails determining the maximum values on boundary directions to determine corners. It does, however, make corners vulnerable to edges. To solve this issue, we need to allow corners to see visual object patterns. Cascade corner pooling searches along a boundary to find a maximum boundary value, find a foremost internal matter, and add the two top values. The corners gain both boundary knowledge and visual patterns of objects by Fig 7(f). | [250] [251] |
| ***g.Adaptive Feature Pooling*** ***MS(2018)*** ***PR(6)*** | Adaptive Feature For each proposal in object detection, pooling gathers features from all levels and fuses them for the next prediction. We assign different feature levels to each submission. RoIAlign is used to pool feature grids from each class, similar to Mask R-CNN. The feature grids from other types are then fused using a fusion process (element-wise max or sum) Fig 7(g). | [252] [253] |
| ***h.Corner Pooling*** ***MS(2018)*** ***PR(6)*** | Corner Pooling is an object detection pooling technique that improves corner localization by encoding explicit prior information. We want to see if a pixel at (i,j) is in the top-left corner. Let $f_t$ and $f_l$ be the vectors at position (i,j) in $f_t$ and $f_l$, respectively, and $f_{tij}$ and $f_{lij}$ be the function maps that are the inputs to the top-left corner pooling sheet. The corner pooling layer max-pools all feature vectors between (i,j) and (i, H) in $f_t$ to a feature vector $t_{ij}$, and max-pools all feature vectors between (i,j) and (W,j) in $f_l$ to a feature vector $l_{ij}$, while using HW feature maps. Finally, it combines $t_{ij}$ and $l_{ij}$ Fig 7(h). | [254] [255] |
| ***i.Generalized Mean Pooling*** ***MS(2000)*** ***PR(2)*** | The comprehensive mean of each channel in a tensor is computed using Generalized Mean Pooling (GeM). Formally: $$\mathrm{e} = [(\frac{1}{\lvert\Omega\rvert}\sum_{u\in\Omega} x_{cu}^{p})^{\frac{1}{p}}]_{c=1,\cdots,C}$$ P>0 denotes a parameter. Setting this exponent to p>1 increases the contrast of the pooled feature map and concentrates attention on the image's most important features. GeM is a simplification of the average pooling (p=1) and spatial max-pooling layer (p=) widely used in classification networks Fig 7(i). | [256] [257] |
| ***j.Local Importance-based Pooling*** ***MS(2019)*** ***PR(1)*** | Local Importance-based Pooling (LIP) is a pooling layer that learns adaptive importance weights based on inputs to improve discriminative features during downsampling. The value function is now not restricted in hand-crafted forms and can understand the criterion for the discriminative features to the use of a learnable network *G* in *F*. In addition, to completely use the function map and avoid the issue of fixed interval sampling, the window size of LIP is limited to not less than stride. In LIP, the critical function is implemented by a small, utterly convolutional network that learns to generate the importance map in an end-to-end manner based on inputs Fig 7(j). | [258] |
| ***k.Hopfield Layer*** ***MS(2020)*** ***PR(1)*** | SoftPool is a method for summing exponentially weighted activations that is quick and efficient. SoftPool holds more detail in downsampled activation maps than a variety of other pooling approaches. Better classification accuracy comes from finer downsampling Fig 7(k). | [259] |
| ***l.RMS Pooling*** ***MS(2000)*** ***0*** | The mean square root for patches of a feature map is calculated and used to construct a downsampled (pooled) feature map in RMS Pooling. After a convolutional sheet, it's typically used Fig 7(l). $$z_j = \sqrt{\frac{1}{M}\sum_{i=1}^{M} uij^2}$$ | |

## 4 RESULTS AND FINDINGS WITH ANALYSIS ON DATASETS

Different models and algorithms performed a vital role in the field of object detection. In the early 1960s, at the Massachusetts Institute of Technology, computer vision originated as an area of Artificial Intelligence (MIT). Owing to the lack of high-performance computers in the 1960s, contributions to this area were minimal. The area received more attention after the invention of supercomputers, and Passement identification, object recognition, scene reconstruction, video tracking, 3D pose estimation, learning, indexing, motion estimation, and image restoration are normally considered sub-domains of computer vision. Object recognition is one of the most common and rapidly evolving sub-domains of computer vision. Because of its use in self-driving vehicles, facial recognition, pediatricians' identification, and surveillance cameras, object recognition has gotten a lot of attention in recent years. The human brain recognizes an object in a fraction of a second, but identifying an object is not a simple task for a computer. It is challenging to define an algorithm that can perform such a task. The system requires many data types, such as images containing one or more objects with corresponding labels. An object recognition system uses prior information from the data to find an object in an image or video belonging to the real world [259]. OD is based on image processing (IP) and Convolutional neural networks (CNN), which are considered enormous branches of computer vision.

### 4.1. Object detection trending application remote sensing image

Object detection, especially in remote sensing images, has seen considerable refinement in recent years. As the quantity and quality of tiny sensing images have improved, this area has seen both opportunities and challenges. Remote sensing images are high-resolution images captured from the planet's surface at various heights and under different lighting conditions for several purposes. By the number of images, the precision of the object detection results can be enhanced. Even though the number of images has risen, it is still not enough to train an effective system with many training images.

Furthermore, the increased number of objects in these images has expanded the scope of the image history, making it more difficult to examine these images accurately. Object identification, on the other hand, is a critical activity for remote sensing images. The task of detecting entire objects in an image, such as roads, water sources, cars, houses, forest fires, and so on, is known as object detection. Detecting all of the objects in the image need to be an efficient and accurate image analysis method.

Remote sensing images are essential and valuable because they monitor and track and change detection and detection shortly. Change identification and environmental monitoring include tracking the development of a population, reducing agricultural lands, reducing forest lands, and the disappearance of water supplies. We're looking at photographs that have been collected over a long period, say over 20 years. We can analyze and make predictions about the condition of any river using these pictures.

Object identification using remote sensing images can be achieved by embedding or simply concatenating a three-step method of candidate area proposals, feature extraction, and final classification of converting these image proposals to labels. The art of drawing regions of interest is known as candidate area proposals. These strategies must focus more on areas of interest, in the sense that there is a possibility of finding an object in that area, rather than regions that make no sense. Feature extraction is the process of extracting each proposal's unique, high-level feature. In general, if the area proposal is better, feature extraction results will be better, leading to more accurate object detection and classification.

**a) Image processing [IP]**

Image processing is a technique for improving the quality of a digital image and extracting valuable information for further processing. It is the discrete representation of a continuous function in a digital image with the help of a two-dimensional array weather pixel is considered a part of the function. Analog image processing and digital image processing are considered as the two main methods of IP. The former deals with two-dimensional signals, while the latter deals with digital pictures. Digital image processing is used in image processing, i.e., color processing and video-based object recognition. Analog image processing is used in hardcopies such as television pictures, photos, and paintings.

***Image Processing Phases***

- **Image Acquisition:** The method of obtaining an image from a source is known as image acquisition. At that stage, the image is usually unprocessed and unaltered.
- **Image Enhancement:** Enhances the quality of an image by enhancing features such as sharpening, improving contrast, and so on.
- **Morphological Processing (MP)**: It is a collection of non-linear operations that only operate on an image's morphological features [489].
- **Image Restoration:** Image restoration reverts an image's quality to its original state by correcting the flaws that caused it to deteriorate.
- **Colour Image Processing:** Color factors such as RGB, HIS, YUV, and CMY describe the image.
- **Image compression:** It is the process of compressing or encoding an image to take up less space than the original.
- **Image segmentation:** It divides an image into distinct regions, allowing perception and analysis to be closely linked within the image.

**b) Convolutional Neural Networks (CNN)**

CNN is a neural network trending in computer vision and solving real-time tracking and detection problems.CNN is the first deep learning technique that effectively understands 2-Dimensional data such as images and voice. CNN provides insight into the quality of photographs and videos; it gives cutting-edge findings in image recognition, tracking, segmentation, scene reconstruction, and human pose prediction, among other things. CNN functions much like an Artificial Neural Network. Yann LeCunn[490] proposed the LeNet model for handwritten digit recognition, the first functional Convolutional Neural Network model. Three primary layers and two sub-layers make it functional:

- Input Layer
- Hidden Layer
  - – Convolution layer
  - – Pooling layer
  - – Fully-Connected layer
- Output Layer

The input layer of CNN takes two-dimensional data as input and passes it on to the subsequent layers.

- ***Convolutional Layer:*** It is considered the heart of CNN and contains many learnable filters. A picture of h*w*r is fed into the convolution layer as input. The letters' h' stands for height, 'w' for width, and 'r' for the color channel. The weights of the filters in the convolution layer remain constant over the function map [260].

Three parameters determine the filter that convolves around the image to find some feature:

– **a. Filter Dimensions:** All of the filters are squares. The ratio between image size and filter size decides the features that can be predicted by filter.

– **b. Padding:** The image pixel's margin is filled with zeros so that no useful information is lost when the filter convolves over the image.

– **c. Stride:** A parameter that specifies how many pixel filters will be horizontally interpreted.

- ***Pooling Layer:*** The pooling layer was implemented to reduce the number of parameters and complicated computations in preparation. Activation maps or function maps are considered as outputs of the convolution sheet. The number of independent activation maps is proportional to the number of filters.
- ***Completely-Connected Layer:*** The fully connected layer receives the pooling layer's output as an input. Many of the previous layers' activation layers are bound to the neurons in the completely corresponding layer.

**Applications Of Convolutional Neural Network (CNN):**

**a)** Karpathy A, A.,[260] argued CNN is used to classify large-scale video. The sports-IM dataset is used to train the CNN architecture. The sports-IM dataset contains 1 million YouTube videos with 487 sports video groups or categories to classify. The architecture was changed to prove generality, and the practical problem of transfer learning with the UCF-101 dataset yielded a 65.4% accuracy.

**b)** In their work, He and Tao [261] use a CNN variant known as 3Dimentional CNN (3D-CNN) to recognize moving objects in images. Since CNN can only learn spatial features from a single frame, the 3D-CNN version retains features directly from the video. Over the multi-view image sequences, a collection of 3D-CNN is implemented in parallel. The spatial-temporal knowledge of the moving object is extracted from each image sequence using 3D-CNN.

There are two types of OD models, and some are two-stage models (TSM), and some are considered one-stage models(OSM). We compared famous object detection algorithms on specific parameters of prime importance and analyzed the performance of those algorithms in Table 8.

Table 8. Object detection models with different parameters and findings

| Approach | R-CNN | FAST R-CNN | FASTER R-CNN | MASK R-CNN | YOLO | SSD |
|---|---|---|---|---|---|---|
| Number of Boxes | 2000 | 2000 | 300 | 300-ResNet, 1000-FPN | 98 | Default set of 8732 |
| Convolutional layer | 5 | 13 | 13 | 13 | 24 | 13 |
| Classifier | SVM | SoftMax | SoftMax | SoftMax | Regression problem | Regression problem |
| Speed up | 1x | 25x | 250x | --- | 500x | --- |
| Frames per Second | --- | 0.5 | 5-17 | 5-17 | 45 | 59 |
| Feature Extractor | VGG-16 | VGG-16 or ResNet | VGG-16 or ZF – 16 | ResNet-FPN | —- | VGG-16 |
| Framework | caffe | caffe | caffe | Caffe | Darknet | caffe |
| Detection stage | 2 | 2 | 2 | 2 | 1 | 1 |
| Loss Function | Log Loss | Smooth L2 | Multitask Loss: (Smooth L1, class Entropy) | Multitask Loss: (Smooth L1, Binary Loss) | Multitask Loss: (Smooth L1) | Multitask Loss: (Smooth L1, Focus loss) |
| Region proposal | Selective search | Selective search | region proposal | CNN | —- | —- |
| Accuracy on the dataset: | | | | | | |
| **i.** PASCAL VOC 2007: | 54 % | 71.8% | 73.2 %70.4% | - | 64.6% | 74.3% |
| **ii.** PASCAL VOC 2012: | 62.3% | 65.7 % | 24.6 % | - | 59.7% | 72.4% |
| **iii.** COCO: | — | — | | 37.1% | 21.6% | 23.2% |

### 4.2. Performance analysis object detection basebone models according to Table 8

Certain conclusions are inferred based on understanding of deep learning algorithms from [260],[268],[271],[273],[304],[305]. By analyzing different contributions Overview about different object detection models presented in Table 8 and advantages/limitation of these methods presented in Table 9. The learning strategy of each model discussed here and discussed in Table 2 is presented in Table 10 concerning literature reference.

- For classification problems, CNN performs admirably.
- R-CNN, YOLO, and SSD are quicker and do better at object detection and recognition.
- For segmentation issues, Mask R-CNN outperforms benchmark and custom datasets.
- When the algorithm's accuracy is considered, Faster R- CNN performs admirably on the benchmark. As compared to other algorithms, the error rate of a faster R-CNN algorithm is extremely low.

Table 9. Different object detection techniques comparative analysis via pros and cons

| No. | Technique | Authors | Year | Advantages | Limitations |
|---|---|---|---|---|---|
| 1. | **Sliding Window[291]** | Viola et al. | 2001 | Simple Easy to implement | Time-consuming |
| 2. | **R-CNN [292]** | Girshick et al. | 2014 | The number of regions proposed is less as compared with the sliding window technique | Multi-stage training expensive training in terms of space and time slow object detection |
| 3. | **OverFeat [292]** | Sermanet et al. | 2014 | High speed than RCNN | Less accurate |
| 4. | **SPP-net [293]** | He et al. | 2014 | Faster than R-CNN, Avoid repeated computation of features. | Reduced accuracy for profound neural network |
| 5. | **MRCNN [295]** | Gidaris et al. | 2015 | Easy to train, Generalize well, Small overhead | Not suited for all kind of real-time applications |
| 6. | **AttentionNet [295]** | Donggeun et al. | 2015 | More accurate detection | Unable to scale to multiple classes, Low recall |
| 7. | **Fast R-CNN[296]** | Girshick et al. | 2015 | High-quality detection than SPPnet and R-CNN | Single-stage training, Slow clustering, Selective search is slow, so still high computation time |
| 8. | **Faster RCNN[297]** | Ren et al. | 2015 | Faster than fast R-CNN | Slow object proposal, Slow implementation than YOLO |
| 9. | **DeepIDNet [298]** | Ouyang et al. | 2015 | Learn deformation of objects with varying size and meaning | Verification issues occur |
| 10. | **YOLO [299]** | Redmon et al. | 2016 | Efficient unified object detector, Extremely fast, Less amount of background errors than fast R-CNN | Can't detect multiple objects within the same grid, Loss inaccuracy rate, Possibility to see one object multiple times., Unable to localize small size objects |
| 11. | **SSD [300]** | Liu et al. | 2016 | Faster than Faster RCNN, Works well for more significant objects | It doesn't generate enough, amount of higher-level features for small objects |
| 12. | **RFCN [301]** | Dai et al. | 2016 | Faster than RCNN with acceptable accuracy, Easier training, Reduced complexity | Need more computation resources |
| 13. | **FPN [302]** | Lin et al. | 2017 | Rich semantics in all levels | Removing top-down connection reduce accuracy |
| 14. | **DeNet [303]** | Tychsen et al. | 2017 | Much faster than RCNN, Predefined anchors not needed | More time spend for generating corners and for evaluating base network |

Table 10. Details of prominent generic object detection architectures.

| Architecture | Proposal generation technique | Learning strategy | Loss function | Softmax layer | End-to-end training | Multi-scale input | Platform |
|---|---|---|---|---|---|---|---|
| RCNN [306] | Selective search | SGD, BP | Hinge Loss | Yes | No | No | Caffe |
| Fast-RCNN [307] | Selective search | SGD | Hinge Loss | Yes | No | No | Caffe |
| Faster-RCNN [308] | Region Proposal Network | SGD | Class Log Loss | Yes | Yes | Yes | Caffe |
| SPP-Net [309] | Edge Boxes | SGD | Hinge Loss | Yes | No | Yes | Caffe |
| R-FCN [310] | Region Proposal Network | SGD | Class Log Loss | No | Yes | Yes | Caffe |
| Mask R-CNN [311] | Region Proposal Network | SGD | Class Log Loss and Semantic Sigmoid Loss | Yes | Yes | Yes | TensorFlow/ Keras |
| FPN [312] | Region Proposal Network | Synchronized SGD | Class Log Loss | Yes | Yes | Yes | TensorFlow |
| SSD [313] | Region Proposal Network | SGD | Class Softmax Loss | No | Yes | No | Caffe |
| YOLO [314] | Region Proposal Network | SGD | Class sum squared error loss | Yes | Yes | No | DarkNet |

### 4.3. Popular datasets and metrics used for general object detection

By object detection, we mean determining whether or not an object belongs to a predefined class. If it does, identifying and locating it in the given image—the abounding box help to describe the location of the object in an image. In different research areas, complex and challenging datasets are considered necessary for comparing different algorithmic approaches and established goals for solutions. Earlier face detection techniques relied on ad hoc databases, but several state-of-the-art face detection datasets were developed recently. Aside from that, datasets for pedestrian detection, face detection, and other problems have been generated. PASCAL VOC [314], ImageNet [315], and MSCOCO [316] are some of the most commonly used object detection benchmarks. In this segment, we'll look at some of the most common datasets and assess their results.

#### *4.3.1. PASCAL VOC dataset:*

From 2005 to 2012, multi-year efforts were devoted to develop a series of benchmark datasets that can be used to detect regular object classes. The PASCAL VOC datasets are made up of 20 different visual object classes spread over 11000 images. Animals, cars, individuals, and domestic objects are the four major divisions of these 20 groups. Furthermore, the semantically related classes of objects, i.e., trucks & buses, sometimes increase detection difficulty levels. In Fig 14, 15, we presented examples from the VOC 2008 and VOC 2012 datasets.

For PASCAL datasets, interpolated average precision (Salton and McGill 1986) was used to evaluate classification and detection accuracy. It was used to chastise detection algorithms for missing object instances, false-positive detections, and repeat detections of a single object instance. After experimenting with mean/average/precision shown in Table 11 for PASCAL VOC 2007 test dataset and Table 12 for PASCAL VOC 2012 test dataset. Eq 1, Eq 2, t is considered the threshold to judge the IoU b/w predicted box & ground truth box. In the VOC metric, the value of t is set to 0.5. $i$ is the index of the i-th image, while $j$ is the index of the j-th object. $N$ is the number of predicted boxes. The indicator function $1[s_{ij} \geq t]= 1$ if $s_{ij} \geq t$ is true, 0 otherwise. According to the threshold criteria, if one detection is matched to a ground truth box, it will be seen as a true positive result. The precision/recall curve is calculated from a method's graded performance for a given task and class. The proportion of all positive examples graded above a given rank is known as recall. Precision is the percentage of positive samples, mAp=mean/average/precision presented in Table 11,12. The final results are the mean average precision for all groups.

$$Recall(t) = \frac{\sum_{ij} 1[s_{ij} \geq t] z_{ij}}{N} \qquad \text{Eq 1.}$$

$$\text{Precision}(t) = \frac{\sum_{ij} 1[s_{ij} \geq t] z_{ij}}{\sum_{ij} 1[s_{ij} \geq t]} \qquad \text{Eq 2.}$$

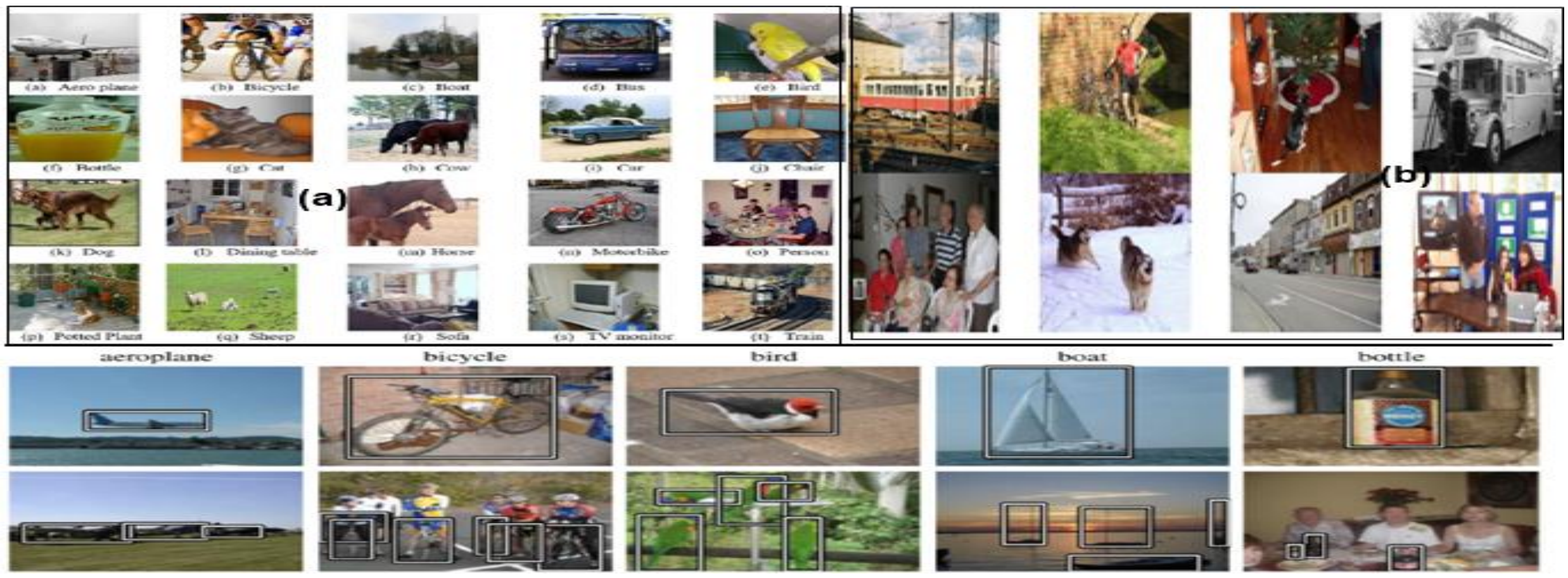

Fig 14. (a)Sample Images from PASCAL VOC 2012 dataset(b) Sample images from PASCAL VOC 2007 dataset(c)Sample images from *PASCAL VOC dataset* [512][513]

Table 11: Performance analysis of some state-of-the-art mean Average Precision (mAP) using PASCAL VOC 2007 test dataset.

| Models | Aeroplan | Bicycle | Bird | Boat | Bottle | Bus | Car | Cat | Chair | Cow | Table | Dog | Horse | mAP | Ref# |
|---|---|---|---|---|---|---|---|---|---|---|---|---|---|---|---|
| R-CNN (Alex) | 68.1 | 72.8 | 56.8 | 43 | 36.8 | 66.3 | 74.2 | 67.6 | 34.4 | 63.5 | 54.5 | 61.2 | 69.1 | 59.10 | [317] |
| R-CNN (VGG) | 73.4 | 77 | 63.4 | 45.4 | 44.6 | 75.1 | 78.1 | 79.8 | 40.5 | 73.7 | 62.2 | 79.4 | 78.1 | 66.98 | [318] |
| SPP-Net | 68.5 | 71.7 | 58.7 | 41.9 | 42.5 | 67.7 | 72.1 | 73.8 | 34.7 | 67 | 63.4 | 66 | 72.5 | 61.58 | [319] |
| OHEM +Fast-RCNN | 80.6 | 85.7 | 79.8 | 69.9 | 60.8 | 88.3 | 87.9 | 89.6 | 59.7 | 85.1 | 76.5 | 87.1 | 87.3 | 79.87 | [322] |
| HyperNet VGG | 84.2 | 78.5 | 73.6 | 55.6 | 53.7 | 78.7 | 79.8 | 87.7 | 49.6 | 74.9 | 52.1 | 86 | 81.7 | 72.01 | [330] |
| Faster R-CNN | 70 | 80.6 | 70.1 | 57.3 | 49.9 | 78.2 | 80.4 | 82 | 52.2 | 75.3 | 67.2 | 80.3 | 79.8 | 71.02 | [320] |
| GCNN | 68.3 | 77.3 | 68.5 | 52.4 | 38.6 | 78.5 | 79.5 | 81 | 47.1 | 73.6 | 64.5 | 77.2 | 80.5 | 68.23 | [323] |
| Bayes | 74.1 | 83.2 | 67 | 50.8 | 51.6 | 76.2 | 81.4 | 77.2 | 48.1 | 78.9 | 65.6 | 77.3 | 78.4 | 69.98 | [324] |
| SDP +CRC | 76.1 | 79.4 | 68.2 | 52.6 | 46 | 78.4 | 78.4 | 81 | 46.7 | 73.5 | 65.3 | 78.6 | 81 | 69.63 | [325] |
| SubCNN | 70.2 | 80.5 | 69.5 | 60.3 | 47.9 | 79 | 78.7 | 84.2 | 48.5 | 73.9 | 63 | 82.7 | 80.6 | 70.69 | [331] |
| StuffNet30 | 72.6 | 81.7 | 70.6 | 60.5 | 53 | 81.5 | 83.7 | 83.9 | 52.2 | 78.9 | 70.7 | 85 | 85.7 | 73.85 | [326] |
| NOC | 76.3 | 81.4 | 74.4 | 61.7 | 60.8 | 84.7 | 78.2 | 82.9 | 53 | 79.2 | 69.2 | 83.2 | 83.2 | 74.48 | [328] |
| MR-CNN + S-CNN | 80.3 | 84.1 | 78.5 | 70.8 | 68.5 | 88 | 85.9 | 87.8 | 60.3 | 85.2 | 73.7 | 87.2 | 86.5 | 79.75 | [327] |
| HyperNet | 77.4 | 83.3 | 75 | 69.1 | 62.4 | 83.1 | 87.4 | 87.4 | 57.1 | 79.8 | 71.4 | 85.1 | 85.1 | 77.20 | [329] |
| SSD300 | 80.9 | 86.3 | 79 | 76.2 | 57.6 | 87.3 | 88.2 | 88.6 | 60.5 | 85.4 | 76.7 | 87.5 | 89.2 | 80.26 | [321] |
| SSD512 | 86.6 | 88.3 | 82.4 | 76 | 66.3 | 88.6 | 88.9 | 89.1 | 65.1 | 88.4 | 73.6 | 86.5 | 88.9 | 82.21 | [321] |

Table 12. Performance analysis of different OD models in mean Average Precision (mAP) using PASCAL VOC 2012 test dataset.

| Models | Aeroplan | Bicycle | Bird | Boat | Bottle | Bus | Car | Cat | Chair | Cow | Table | Dog | Horse | mAP | Ref# |
|---|---|---|---|---|---|---|---|---|---|---|---|---|---|---|---|
| R-CNN (Alex) | 71.8 | 65.8 | 52 | 34.1 | 32.6 | 59.6 | 60 | 69.8 | 27.6 | 52 | 41.7 | 69.6 | 61.3 | 53.68 | [317] |
| R-CNN (VGG) | 79.6 | 72.7 | 61.9 | 41.2 | 41.9 | 65.9 | 66.4 | 84.6 | 38.5 | 67.2 | 46.7 | 82 | 74.8 | 63.34 | [318] |
| Fast-RCNN | 82.3 | 78.4 | 70.8 | 52.3 | 38.7 | 77.8 | 71.6 | 89.3 | 44.2 | 73 | 55 | 87.5 | 80.5 | 69.34 | [318] |
| OHEM +Fast-RCNN] | 90.1 | 87.4 | 79.9 | 65.8 | 66.3 | 86.1 | 85 | 92.9 | 62.4 | 83.4 | 69.5 | 90.6 | 88.9 | 80.64 | [332] |
| Faster R-CNN | 84.9 | 79.8 | 74.3 | 53.9 | 49.8 | 77.5 | 75.9 | 88.5 | 45.6 | 77.1 | 55.3 | 86.9 | 81.7 | 71.63 | [320] |
| StuffNet30 | 83 | 76.9 | 71.2 | 51.6 | 50.1 | 76.4 | 75.7 | 87.8 | 48.3 | 74.8 | 55.7 | 85.7 | 81.2 | 70.65 | [326] |
| NOC | 82.8 | 79 | 71.6 | 52.3 | 53.7 | 74.1 | 69 | 84.9 | 46.9 | 74.3 | 53.1 | 85 | 81.3 | 69.85 | [328] |
| MR-CNN + S-CNN | 85.5 | 82.9 | 76.6 | 57.8 | 62.7 | 79.4 | 77.2 | 86.6 | 55 | 79.1 | 62.2 | 87 | 83.4 | 75.03 | [327] |
| HyperNet | 84.2 | 78.5 | 73.6 | 55.6 | 53.7 | 78.7 | 79.8 | 87.7 | 49.6 | 74.9 | 52.1 | 86 | 81.7 | 72.01 | [329] |
| ION | 87.5 | 84.7 | 76.8 | 63.8 | 58.3 | 82.6 | 79 | 90.9 | 57.8 | 82 | 64.7 | 88.9 | 86.5 | 77.19 | [333] |
| SSD512 | 91.4 | 88.6 | 82.6 | 71.4 | 63.1 | 87.4 | 88.1 | 93.9 | 66.9 | 86.6 | 66.3 | 92 | 91.7 | 82.31 | [321] |
| SSD300 | 91 | 86 | 78.1 | 65 | 55.4 | 84.9 | 84 | 93.4 | 62.1 | 83.6 | 67.3 | 91.3 | 88.9 | 79.31 | [321] |
| YOLO | 77 | 67.2 | 57.7 | 38.3 | 22.7 | 68.3 | 55.9 | 81.4 | 36.2 | 60.8 | 48.5 | 77.2 | 72.3 | 58.73 | [313] |
| YOLO + Fast R-CNN | 83.4 | 78.5 | 73.5 | 55.8 | 43.4 | 79.1 | 73.1 | 89.4 | 49.4 | 75.5 | 57 | 87.5 | 80.9 | 71.27 | [313] |

### *4.3.2. MS COCO BENCHMARK*

There are 91 common object groups in the Microsoft common objects in context (MS COCO) dataset [362] for detecting and segmenting objects encountered in daily life in their natural environments, with 82 of them having more than 5,000 named instances. These categories comprise the PASCAL VOC dataset's 20 categories. In total, there are 2,500,000 named instances in 328,000 images in the dataset. MS COCO dataset also considers various points of view, and all of the objects are found in natural settings, providing us with a wealth of contextual knowledge.
COCO has fewer categories but more instances per category than the common ImageNet dataset.
The dataset has slightly more instances per category (27k on average) than the PASCAL VOC datasets [361] (about 10 times less than the MS COCO dataset) and the ImageNet object detection dataset (1k) [360]. Compared to PASCAL VOC (2.3) and ImageNet, MS COCO has significantly more object instances per image (7.7). (3.0). Furthermore, the MS COCO dataset has 3.5 categories per image, compared to 1.4 and 1.7 for PASCAL and ImageNet, respectively. Furthermore, 10% of images in MS COCO have only one category, while more than 60% of images in ImageNet and PASCAL VOC have only one object category. Small items, as we all know, need more contextual thinking to understand. Contextual details abound in the MS COCO dataset's images. The largest class is ''person," with nearly 800,000 instances, while the smallest class is ''hair driver," with just 600 instances in the entire dataset. Another limited class is "hairbrush," which has nearly 800 members. Except for 20 classes with a large or small number of instances, the remaining 71 types have approximately the same number of instances. In the first two lines of Fig. 15, three standard image categories from the MS COCO dataset are shown. The MS COCO metric is used to judge the success of detections strictly and thoughtfully. The threshold in PASCAL VOC is set to a single value, 0.5, but the mean average precision in MS COCO is calculated using a range of [0.5,0.95] with an interval of 0.05, or 10 values. Besides that, the exceptional average precision for small, medium and large objects is measured separately to assess the detector's ability to detect targets of various sizes. Results after performing average- precision is shown in Table 13 for single-stage and two-stage object detectors.

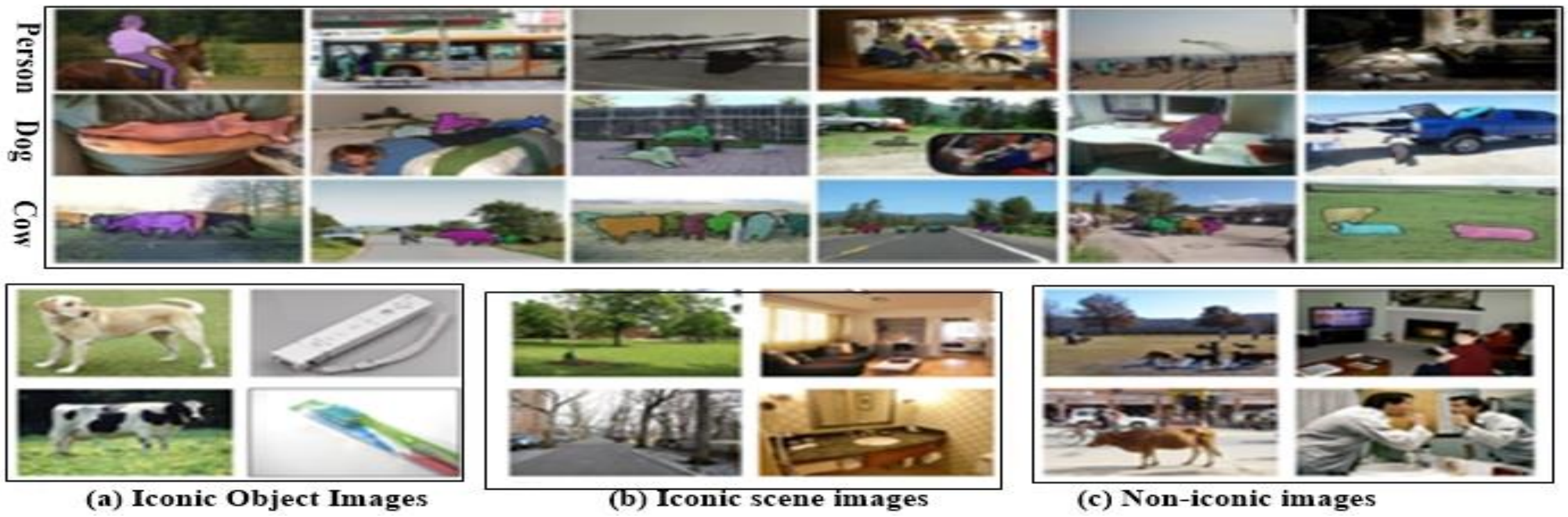

Fig 15. Sample annotated images from the MS-COCO dataset [514][515]

Table 13. Comparison of detection performance of some state-of-the-art techniques on MSCOCO test-dev dataset.

| Model | Backbone architecture | Year of introduction | AP | $AP_{50}$ | $AP_{75}$ | $AP_S$ | $AP_M$ | $AP_L$ | |
|---|---|---|---|---|---|---|---|---|---|
| **Single-Stage-Object-Detectors** | | | | | | | | | |
| SSD512 | VGG-16 | 2016 | 28.8 | 48.5 | 30.3 | 10.9 | 31.8 | 43.5 | [340] |
| SSD513 | ResNet-101 | 2017 | 31.2 | 50.4 | 33.3 | 10.2 | 34.5 | 49.8 | [336] |
| DSSD513 | ResNet-101 | 2017 | 33.2 | 53.3 | 35.2 | 13.0 | 35.4 | 51.1 | [336] |
| STDN513 | DenseNet-169 | 2018 | 31.8 | 51.0 | 33.6 | 14.4 | 36.1 | 43.4 | [334] |
| CornerNet511 | Hourglass-169 | 2018 | 40.5 | 56.5 | 43.1 | 19.4 | 42.7 | 53.9 | [338] |
| CornerNet511 | Hourglass-169 | 2019 | 44.9 | 62.4 | 48.1 | 25.6 | 47.4 | 57.4 | [339] |
| GHM SSD | ResNeXt-101 | 2018 | 41.6 | 62.8 | 44.2 | 22.3 | 45.1 | 55.3 | [335] |
| FPN-Reconfig | ResNeXt-101 | 2018 | 34.6 | 54.3 | 37.3 | NA | NA | NA | [337] |
| FCOS | ResNeXt-101 | 2019 | 42.1 | 62.1 | 45.2 | 25.6 | 44.9 | 52.0 | [341] |
| FSAF | ResNeXt-101 | 2019 | 42.9 | 63.8 | 46.3 | 26.6 | 46.2 | 52.7 | [343] |
| ExtremeNet | Hourglass-104 | 2019 | 40.2 | 55.5 | 43.2 | 20.4 | 43.2 | 53.1 | [342] |
| M2Det800 | VGG-16 | 2019 | 41.0 | 59.7 | 45.0 | 22.1 | 46.5 | 53.8 | [344] |
| RefineDet512 | ResNet-101 | 2018 | 36.4 | 57.5 | 39.5 | 16.6 | 39.9 | 51.4 | [345] |
| YOLOv2 | DarkNet-19 | 2017 | 21.6 | 44.0 | 19.2 | 5.0 | 22.4 | 35.5 | [358] |
| **Two-Stage-Object-Detectors** | | | | | | | | | |
| Fast R-CNN | VGG-16 | 2015 | 19.7 | 35.9 | NA | NA | NA | NA | [318] |
| Faster R-CNN | VGG-16 | 2015 | 21.9 | 42.7 | NA | NA | NA | NA | [320] |
| Faster R-CNN w FPN | ResNet-101 | 2016 | 36.2 | 59.1 | 39.0 | 18.2 | 39.0 | 48.2 | [352] |
| Faster R-CNN by G-RMI | Inception-ResNet-v2 | 2017 | 34.7 | 55.5 | 36.7 | 13.5 | 38.1 | 52.0 | [349] |
| OHEM | VGG-16 | 2016 | 22.6 | 42.5 | 22.2 | 5.0 | 23.7 | 37.9 | [332] |
| ION | VGG-16 | 2016 | 23.6 | 43.2 | 23.6 | 6.4 | 24.1 | 38.3 | [333] |
| R-FCN | ResNet-101 | 2016 | 29.9 | 51.9 | NA | 10.8 | 32.8 | 45.0 | [353] |
| CoupleNet | ResNet-101 | 2017 | 34.4 | 54.8 | 37.2 | 13.4 | 38.1 | 50.8 | [57] |
| Deformable R-FCN | Aligned-Inception-ResNet | 2017 | 37.5 | 58.0 | 40.8 | 19.4 | 40.1 | 52.5 | [353] |
| DeNet-101 | ResNet-101 | 2017 | 33.8 | 53.4 | 36.1 | 12.3 | 36.1 | 50.8 | [354] |
| Mask-RCNN | ResNeXt-101 | 2017 | 39.8 | 62.3 | 43.4 | 22.1 | 43.2 | 51.2 | [310] |
| Fitness-NMS | ResNet-101 | 2017 | 41.8 | 60.9 | 44.9 | 21.5 | 45.0 | 57.5 | [356] |
| Relation Net | ResNet-101 | 2018 | 39.0 | 58.6 | 42.9 | NA | NA | NA | [355] |
| DeepRegionlets | ResNet-101 | 2018 | 39.3 | 59.8 | NA | 21.7 | 43.7 | 50.9 | [346] |
| C-Mask RCNN | ResNet-101 | 2018 | 42.0 | 62.9 | 46.4 | 23.4 | 44.7 | 53.8 | [347] |
| DCN + R-CNN | ResNet-101 + ResNet-152 | 2018 | 42.6 | 65.3 | 46.5 | 26.4 | 46.1 | 56.4 | [351] |
| Cascade R-CNN | ResNet-101 | 2018 | 42.8 | 62.1 | 46.3 | 23.7 | 45.5 | 55.2 | [348] |
| Grid R-CNN | ResNeXt-101 | 2019 | 43.2 | 63.0 | 46.6 | 25.1 | 46.5 | 55.2 | [350] |
| DCN-v2 [121] | ResNet-101 | 2019 | 44.8 | 66.3 | 48.8 | 24.4 | 48.1 | 59.6 | [359] |
| TridentNet [195] | ResNet-101 | 2019 | 42.7 | 63.6 | 46.5 | 23.9 | 46.6 | 56.6 | [349] |

### *4.3.3. ImageNet*

A move forward in vision activities and realistic implementations can be encouraged by challenging datasets. The ImageNet dataset [360] is another significant large-scale benchmark dataset. The ILSVRC task of object detection assesses an algorithm's ability to identify all target objects' instances in an image. There are approximately 450k training images, 20k validation images, and 40k test images in ILSVRC2014. Small object deviations of a few pixels, on the other hand, will be inappropriate according to this threshold. ImageNet employs a loosen threshold measured as follows: where w and h denote the width and height of a ground fact box, as necessary, respectively. This value allows the annotation to stretch up to 5 pixels in each direction around the object on average—table 14 present different datasets with training and validation subset values of the number of images.

Table 14. Some well-known object detection datasets and their statistics

| **Dataset** | **Subset of Training** | | **Subset of validation** | | **Subset of Travel** | | **Test Subset** | |
|---|---|---|---|---|---|---|---|---|
| | images | objects | images | objects | images | objects | images | objects |
| VOC-2007 | 2,501 | 6,301 | 2,510 | 6,307 | 5,011 | 12,608 | 4,952 | 14,976 |
| | | | | | 11,540 | | | |
| VOC-2012 | 5,717 | 13,609 | 5,823 | 13,841 | | 27,450 | 10,991 | - |
| ILSVRC-2014 | 456,567 | 478,807 | 20,121 | 55,502 | 476,688 | 534,309 | 40,152 | - |
| ILSVRC-2017 | 456,567 | 478,807 | 20,121 | 55,502 | 476,688 | 534,309 | 65,500 | - |
| MS-COCO-2015 | 82,783 | 604,907 | 40,504 | 291,875 | 123,287 | 896,782 | 81,434 | - |
| MS-COCO-2018 | 118,287 | 860,001 | 5,000 | 36,781 | 123,287 | 896,782 | 40,670 | - |
| OID-2018 | 1,743,042 | 14,610,229 | 41,620 | 204,621 | 1,784,662 | 14,814,850 | 125,436 | 625,282 |

### *4.3.4. VisDrone2018 BENCHMARK*

VisDrone2018 [364], large-scale visual object detection and tracking benchmark dataset were released last year. It consists of images and videos captured by drones. The purpose of this dataset is to improve visual comprehension tasks on the drone platform. The images and video sequences in the benchmark were captured in various urban/suburban areas of 14 different cities from north to south China. VisDrone2018, in particular, is made up of 263 video clips and 10,209 images (no overlap with video clips) with rich annotations such as object bounding boxes, object categories, occlusion, truncation ratios, and so on. This benchmark has over 2.5 million annotated instances spread across 179,264 images/video frames.

As the most extensive such dataset ever published, the benchmark allows for comprehensive evaluation and investigation of visual analysis algorithms on the drone platform. VisDrone2018 contains many small objects, such as dense cars, pedestrians, and bicycles, making detection difficult in specific categories. Furthermore, 82.4 percent of the images in the training set have more than 20 objects per image, and the average number of objects per image is 54 in the 6471 images in the training set. Because this dataset contains night scenes, the brightness of these images is lower than that of daytime images, complicating the detection of small and dense objects, as shown in Fig. 15. The MS COCO metric is used in this dataset.

### 4.3.5. OPEN IMAGES V5

Open Images [13] is a dataset with image-level labels, object bounding boxes, object segmentation masks, and visual relationships annotated on 9.2 million images. Available Images V5 has a total of 16 million bounding boxes for 600 object groups on 1.9 million images, making it the largest dataset with object position annotations currently available. To begin, competent annotators (Google-internal annotators) drew the boxes in this dataset by hand to ensure accuracy and consistency. Second, the images in it are incredibly varied, with most of them containing complex scenes of multiple objects (8.3 per image on average). Third, this dataset includes visual relationship annotations that show pairs of objects in specific relationships (for example, "woman playing the guitar" and "beer on the table"). There are 329 relationship triplets in total, with 391,073 samples. V5 also includes segmentation masks for 2.8 million object instances across 350 classes. The out-line of artefacts is marked by segmentation masks, which characterizes their spatial scale to a much higher level of detail. Finally, 36.5 million image-level labels spanning 19,969 groups have been added to the dataset. Kuznetsova et al. [13] suggest modifications based on the PASCAL VOC 2012 mAP evaluation metric to consider several essential aspects of the Open Images Dataset thoroughly. To avoid being mistakenly counted as false negatives, the unannotated classes are ignored for equal evaluation. Second, if an object belongs to both a class and a subclass, an object detection model can produce detection results for both classes. In that class, the absence of one of these classes will be called a false negative. Third, there are group-of-boxes in the Open Images Dataset that contain a group of (more than one that is impeding or physically touching) object instances but no single object localization inside them. A detection within a group of the box will be counted as a true positive if the intersection of the detection and the box separated by the detection region is more significant than 0.5. Only one valid true positive is counted when several correct detections are made within the same group of boxes Eq 3.

$$t = \min\left(0.5, \frac{wh}{(w+10)(h+10)}\right. \qquad \text{Eq 3.}$$

## 4.4. Some more famous datasets result for Tiny object detection

Datasets always play an important role in every innovative concept evaluation for object detection because they allow for standard comparison of competing algorithms and solution goals. PASCAL-VOC [367], MS-COCO [365], FlickrLogos [368,369], KITTI [370], SUN [371], Tsinghua-Tencent 100 K (TT100K) [372], Caltech [373], and other well-known datasets have been published in recent years. MS-COCO and PASCAL-VOC are two programs for detecting generic objects. Caltech and KITTI are used to detect pedestrians. Furthermore, FlickrLogos, TT100K, and SUN are used to detect logos, traffic signals, and scenes, respectively. The small object dataset (SOD) [366] is specifically designed to address small object detection issues. See Table 15 for more information.

Table 15, Some famous datasets with detailed information

| Rank | Dataset | Special useful purpose | Explanation | Source | Published in | Ref# |
|---|---|---|---|---|---|---|
| 1 | PASCAL-VOC | Most iconic object detection | It is considered the most famous object detection dataset widely. PASCAL-VOC is used in papers with two different versions: VOC2007 and VOC2012. There are 2501 training images, 2510 validation images and 5011 testing images in VOC2007. VOC2012, on the other hand, has 5717 training images, 5823 validation images, and 10,991 test images. They're both 20-category mid-scale datasets for object detection. | http://host.robots.ox.ac. UK/pascal/VOC/. | IJCV | [367] |
| 2 | MS-COCO | Most popular and challenging object detection datasets | It is one of the most popular now and considered challenging object detection datasets. It has 164,000 images and 897,000 labelled objects divided into 80 categories. It consists of three image splits training, validation, and testing. The training, validation, and testing sets each have 118,287, 5000, and 40,670 photos, respectively. MS-object COCO's distribution is more closer to real-world settings. The MS-COCO testing set annotation information is not accessible. | http://cocodataset. Org. | ECCV | [365] |
| 3 | KITTI | Traffic scene analysis/detection | It is a well-known dataset for traffic scene analysis. It contains 7481 labelled images and another 7518 images for testing. There are 100,000 instances of pedestrians. With around 6000 identities and one person average on per image. In KITTI, there are two subclasses of people: pedestrians and cyclists. Based on the extent to which the items are occluded and truncated, the object labels are categorized into easy, moderate, and tough categories. | http://www. cvlibs.net/datasets/kitti/index.php . | CVPR | [370] |
| 4 | Caltech | Pedestrian detection datasets | It's internationally regarded as one of the most popular and challenging datasets for pedestrian identification. In the training and testing sets, there are 192,000 and 155,000 pedestrian incidents, respectively. | http://www.vision.caltech. edu/Image_Datasets/CaltechPedestrians/. | PAMI | [373] |
| 5 | Flickr Logos | FlickrLogos-47 uses the same image corpus | FlickrLogos-32 and FlickrLogos-47 are two versions of FlickrLogos that are often used in papers. FlickrLogos-47 uses the same image corpus as FlickrLogos-32, but it adds new classes and annotations to missing object instances. There are 833 training images and 1402 testing images in FlickrLogos-47. | http://www.multimedia-computing.de/flickrlogos/. | ICMR | [368,369] |
| 6 | SUN | Scene recognition & object detection | It has approximately 132,000 images divided into 908 different scene types. SUN397 and SUN2012 are the most commonly used forms of SUN in the literature. The first is used to recognize scenes, while the second is used to detect objects. It has approximately 132,000 images divided into 908 different scene types. SUN397 and SUN2012 are the most commonly used forms of SUN in the literature. The first is used to recognize scenes, while the second is used to detect objects. | http://groups.csail.mit.edu/vision/SUN/. | IJCV | [371] |
| 7 | TT100K | Traffic sign detection | With 100,000 photos and 30,000 traffic sign instances of 128 classes, it is the most significant traffic sign detection collection to date. The images have a resolution of up to 2048 by 2048 pixels, though common traffic sign examples are fewer than 32 x 32 pixels. A class label, bounding box and pixel mask are all added to each instance. It show features of many small items, a lot of lighting and many scale variations. There are 45 classes in all, each having at least 100 instances. | http://cg.cs. tsinghua.edu.cn/traffic%2Dsign/. | CVPR | [372] |
| 8 | SOD | | A subset of images from both the MS-COCO and SUN datasets are used to create the small object dataset (SOD). SOD has roughly 8393 object instances spread across 4925 images in 10 categories. The object categories chosen are "mouse," "telephone," "switch," "outlet," "clock," "toilet paper," "tissue box," "faucet," "plate," and "jar." SOD's object instances all are small. Please see the following website for more details. | http://www.merl.com. | ACCV | [366] |

#### 4.4.1. Evaluation of results by applying different techniques for tiny object detection

We have thoroughly summarized the methods for detecting small objects from five perspectives. Several approaches support object detection, i.e. Multi-scale feature learning, data augmentation, training strategy, context-based detection, and GAN-based detection, each of them has a unique role and strategy.

Tables 15 show the detection results of these cutting-edge algorithms on the MS-COCO based on our taxonomy of small object detection algorithms. It is examined MS-COCO shows better results by applying object detection approaches than other test sets. This work because the MS-COCO dataset has more objects per image than the PASCAL-VOC dataset. Compared to the PASCAL-VOC dataset, most MS COCO datasets are small with large-scale ranges, resulting in poor detection results.

As shown in Table 15, large objects have the best detection performance among different-sized objects, i.e., large, medium, and small objects (presented in columns 1, 2, and 3), while small objects have the worst. This dataset also demonstrates the difficulty of detecting small objects SNIPER [180] improves the SNIP [172] from 45.7 percent AP to 46.1 percent AP and the SNIP on small objects from 29.3 percent to 29.6 percent. On the test-devset, SNIP with multi-scale feature learning and training strategy achieves 45.7 percent AP, outperforming other algorithms (except SNIPER) by a wide margin. Meanwhile, SNIP's AP on small objects remains higher than other methods (except SNIPER).

Furthermore, FPN variants such as RefineDet512++ [169] and M2Det800++ [181] achieve more than 25% AP on small object detection. R-FCN++ [171] and MTGAN [177] detect small objects using global context information and multi-task GANs, respectively, and achieve over 24 percent AP on small objects.

Furthermore, ION [157] used VGG16 as the backbone network and achieved 79.2 percent mAP and 76.4 percent mAP, which is better than other algorithms that use VGG16. It is worth noting that ION achieves good performance by leveraging integrated features and global context information.

The following observations can be drawn from the above comparative analysis of Table 16.

- ➢ On the MS-COCO, multi-scale feature learning, data augmentation, training strategy, context-based detection, and GAN-based detection methods can improve the detection performance of small objects.
- ➢ Using different input resolutions, such as SSD300 [158] and SSD512 [158], DSSD321 [159] and DSSD513 [159], affects detection performance of small objects. Higher input resolution may yield better detection results than lower input resolution.
- ➢ More efficient backbone CNN models, when properly combined, will improve small object detection efficiency.
- ➢ The detection performance of SNIP, ION on the MS-COCO dataset demonstrates that combining several small object detection methods can significantly boost small object detection performance.
- ➢ For deep learning-based detection models (SSD512 with '07', '07 + 12', and '07 + 12 + COCO'), data augmentation is critical for obtaining robust features also be presented if we do further analysis.

Due to keeping in mind the limit of paper length, we shortened our experimental observations if the same future type of experiments can be performed on other datasets (PASCAL-VOC, FlickrLogos, KITTI, SUN, Tsinghua-Tencent 100 K (TT100K), Caltech ) as well for comparative analysis.

Table 16. Detection results on the MS-COCO test-dev dataset of some typical methods. "++" denotes applying inference strategy such as multi-scale test, horizontal flip, etc. (in %).

| Type | | Method | Backbone | AP | $AP_{50}$ | $AP_{75}$ | $AP_S$ | $AP_M$ | $AP_L$ | Ref# |
|---|---|---|---|---|---|---|---|---|---|---|
| Multi-scale feature learning | Featurized image pyramids | SNIP | DPN98 | 45.7 | 67.3 | 51.1 | 29.3 | 48.8 | 57.1 | [172] |
| | Single feature map | Faster R-CNN | VGG16 | 21.9 | 42.7 | – | – | – | – | [148] |
| | Pyramidal feature hierarchy | SSD300 | VGG16 | 23.2 | 41.2 | 23.4 | 5.3 | 23.2 | 39.6 | [158] |
| | | SSD512 | VGG16 | 26.8 | 46.5 | 27.8 | 9.0 | 28.9 | 41.9 | [158] |
| | Integrated features | ION | VGG16 | 24.6 | 46.3 | 23.3 | 7.4 | 26.2 | 38.8 | [157] |
| | Feature pyramid network | FPN | ResNet101 | 36.2 | 59.1 | 39.0 | 18.2 | 39.0 | 48.2 | [161] |
| | Variants of FPN | RefineDet512 | ResNet101 | 36.4 | 57.5 | 39.5 | 16.6 | 39.9 | 51.4 | [169] |
| | (including feature fusion and feature pyramid generation, | RefineDet512 ++ | ResNet101 | 41.8 | 62.9 | 45.7 | 25.6 | 45.1 | 54.1 | [169] |
| | multi-scaled fusion module, etc.) | M2Det800 | VGG16 | 41.0 | 59.7 | 45.0 | 22.1 | 46.5 | 53.8 | [181] |
| | | M2Det800++ | VGG16 | 44.2 | 64.6 | 49.3 | 29.2 | 47.9 | 55.1 | [181] |
| | | FSSD300 | VGG16 | 27.1 | 47.7 | 27.8 | 8.7 | 29.2 | 42.2 | [166] |
| | | FSSD512 | VGG16 | 31.8 | 52.8 | 33.5 | 14.2 | 35.1 | 45.0 | [166] |
| | | DSSD321 | ResNet101 | 28.0 | 46.1 | 29.2 | 7.4 | 28.1 | 47.6 | [159] |
| | | DSSD513 | ResNet101 | 33.2 | 53.3 | 35.2 | 13.0 | 35.4 | 51.1 | [159] |
| | | MDSSD300 | VGG16 | 26.8 | 46.0 | 27.7 | 10.8 | – | – | [182] |
| | | MDSSD512 | VGG16 | 30.1 | 50.5 | 31.4 | 13.9 | – | – | [182] |
| Data augmentation | | Augmentation | ResNet50 | 30.4 | – | – | 17.9 | 32.9 | 38.6 | [183] |
| Training strategy | | SNIP | DPN98 | 45.7 | 67.3 | 51.1 | 29.3 | 48.8 | 57.1 | [172] |
| | | SNIPER | ResNet101 | 46.1 | 67.0 | 51.6 | 29.6 | 48.9 | 58.1 | [180] |
| | | SAN | R-FCN | 36.3 | 59.6 | – | 16.7 | 40.5 | 55.5 | [176] |
| Context-based detection | Local context | MPNet | ResNet | 33.2 | 51.9 | 36.3 | 13.6 | 37.2 | 47.8 | [153] |
| | | CoupleNet | ResNet101 | 33.1 | 53.5 | 35.4 | 11.6 | 36.3 | 50.1 | [164] |
| | | CoupleNet++ | ResNet101 | 34.4 | 54.8 | 37.2 | 13.4 | 38.1 | 50.8 | [164] |
| | Global context | ION | VGG16 | 24.6 | 46.3 | 23.3 | 7.4 | 26.2 | 38.8 | [157] |
| | | R-FCN++ | R-FCN | 42.3 | 63.8 | – | 25.2 | 46.1 | 54.2 | [171] |
| | | ORN | ResNet50 | 30.5 | 50.2 | 32.4 | – | – | – | [173] |
| | | SIN | VGG16 | 23.2 | 44.5 | 22.0 | 7.3 | 24.5 | 36.3 | [174] |
| GAN-based detection | | MTGAN | ResNet101 | 41.4 | 63.2 | 45.4 | 24.7 | 44.2 | 52.6 | [177] |

## 5 CONCLUSION AND CHALLENGING FUTURE DIRECTIONS

### 5.1. CONCLUSION

This survey paper mainly contributed to three main areas:

1. Computer vision and general object detection
2. Deep domain adaptive object detection and deep learning
3. Tiny object detection as a trending research problem

We presented comparative and computational results with the help of the literature and using different famous datasets. Instead of detailed information, we focused on the main description and showed results in tabular form so that readers gain more knowledge in short paper length requirements.

Deep learning-based object detection techniques have intentionally become a highly researched area due to their powerful learning capabilities and utility in interacting with occlusions, scale shifts, and background exchanges. A detailed survey of the most recent developments in deep learning-based visual object detection is presented in this paper. The study begins with a survey of recent literature, followed by examining traditional and modern detectors. After that, a thorough examination of backbone architectures is carried out and presented a systematic analysis of effective learning strategies. We analyzed different problems in tiny and general object detection and provided solution-based techniques with literature references.

Five deep domain adaptive object detection (DDAOD) approaches have also been examined in this paper. The techniques that we considered for discussion are classified using the five main categories, and after implementation, object detection methods on datasets results are presented in tables. The results of various domain adaptive object detection tasks are compared and used detection models inside. It can imagine that hybrid methods received the most top positions, adversarial-based methods received the most and received the minor top positions. It has been demonstrated that adversarial training with more adaptation mechanisms is more successful. Although several DDAOD methods have been proposed in recent years, there is still a significant gap between achieved output and the oracle results of a detector trained in labeled data targets.

The identification of small objects is a challenging problem in computer vision. This paper compares and analyzes the latest classic small object detection algorithms on some familiar object detection datasets, such as PASCAL-VOC, MS-COCO, KITTI, and TT100K, and examines tiny object detection methods deep learning from five dimensions. The results from a few of these datasets show that multi-scale feature learning, data augmentation, training strategy, context-based detection, and GAN-based detection methods can all improve small object detection efficiency.

Object detection technology with deep learning became advanced quickly due to its efficient computing equipment upgrade power. The need for high-precision real-time systems is becoming urgent needs to deploy on more precise applications. Researchers have developed several directions, including building new architecture, extracting rich features, exploiting good representations, improving processing speed, training from scratch, anchor-free methods, solving sophisticated scene issues (small objects, occluded objects) combining one-stage and multi-stage detectors. The application of object detection is steadily expanding, applications of robust object detectors are increasing in the defense, military, transportation, medical, and life fields. In addition, there are several divisions in the detection domain.

Finally, to gain a detailed understanding of the object detection landscape, some standard datasets and benchmarks for visual object detection are addressed with some application areas. Object detection applications are growing dramatically as optical object detectors become more effective in defense, transportation, and the military.

Although the promising progress of this field has recently been achieved, the gap between modern and human performance is still enormous. There is still need a lot of research in the following aspects discussed in section 6.2.

#### *5.1.1. Summery about paper overview*

Deep learning currently attained attention due to its trending importance for humanoid tracking/detection applications. This paper offered a comprehensive analysis of previous and current approaches that helped in object detection. In the introduction part, section 1, Table 2 is a review of more than 125 papers. By looking at it, anyone can get a complete idea of the deep learning-based object detection field. We explained our concept with the support of prior surveys and explained the significant role of old work in computer vision. In section 2, We provided a solution for object detection with a strategic analysis of different approaches. In the 2.1 section, we explained deep domain adaptive object detection DDAOD approaches such as discrepancy-based DDAOD, adversarial-based DDAOD, reconstruction-based DDAOD, hybrid DDAOD, and some other DDAOD approaches. These OD methodologies are mapped with OD algorithms, i.e., R-CNN, Fast R-CNN, Faster R-CNN, SDD, and YOLO. We presented these analyses with the help of different datasets, and analysis results can be seen in Table 3(3.1,3.2,3.3,3.4 and 3.5).

In section 2.2, we addressed tiny object detection (TOD) with strategic and historical analysis. Table 4 presented TOD state-of-the-art from 2014 to 2020 and presented highly reputed papers with their proposed methods and chronological order methodologies. In Table 5, we explained the taxonomy of methods for OD with supporting sub-categories and processes. After that, OD-related approaches, i.e., multi-scale feature learning, data augmentation, training strategy, context-based detection, GAN-based detection, and general state-of-the-art techniques explained in section 2.3 with Table 6, presented face Detection strategies with one/two-stage models. In section 3 explains object detection methods and types of these methods, focusing on convolution, convolution neural networks, and pooling operations with the combined picture of many techniques at the same place. In section 4 explained the above OD algorithms with a comparative outlook in sections 2,3,5, Table 2. This section gave a short brief on convolutional neural networks (CNN) and deep learning (DL), as these algorithms (R-CNN, Fast R-CNN, Faster R-CNN, YOLO, SSD), from many years these OD models working as the backbone for the latest detection-based trends. After that, we offered a comparative analysis of these methods' working thresholds in Table 7. Then we highlighted the basic description & main applications of these models in OD with literature reference Table 8 and essential parameters of OD. In Table 9, the advantages and limitations of these OD models are presented comparatively. For compative analysis we used some fundamental OD datasets. After performing different OD models with famous datasets, i.e., VOC PASCAL, COCO, and ImageNet datasets, after conducting experiments on datasets with OD approaches/ Methods/Models final results presented in Table 10-16.In section 5, we offered a comprehensive conclusion and future direction. In short, we can say our survey paper is unique in deep learning according to discussed domains and comparative analysis point of view.

### 5.2. Overview of object detection field, challenges and future directions

This section reviewed important trending detection research directions with literature references and some of them presented in Table 2. By examining more than 500 latest trending research contributions, we found some future directions that can be helpful for more research.

#### 5.2.1. *Future directions for general object detection*

- **Unsupervised Detection Objects detection:** Automated annotation technologies are exciting and promising for unattended object detection to eliminate manual annotation. Unmonitored object detection is a future direction of research for sharp detection tasks.
- **Real-Time Detection Remote sensing:** Remote sensing images are used in both military and agricultural areas. The fast development of these fields will enhance automated detection models related to hardware devices.
- **Weakly supervised object detection:** Weakly supervised object detection patterns are used to detect many unannotated counterparts through a small set of fully annotated images. That is a significant problem for future research. Large quantities of annotated and labeled images are used to effectively train a network for achieving high efficiency, using target objects and bounding boxes.
- **Video Object Detection:** As we know, video object detection suffers from motion objective ambiguities, tiny target objects, truncations, and occlusions, and obtaining high accuracy and efficiency is highly considered a difficult task. Research on motion-based objectives and multifaceted data sources like video sequences are now considered one of the more promising areas for future research.
- **Multi-domain object detection:** We know that field-specific detectors always achieve higher detection precision on a predetermined dataset. Therefore, the future primarily develops a universal object detector that can detect multi-domain objects without previous knowledge.
- **Salient Object Detection:** This section is designed to emphasize the virtual object in an image. A wide range of object detection applications in different areas is used for the detection of Salient objects. The critical salient object regions can help detect the objects accurately in a continuous scene or video sequence in each frame. Salt detection guided objects can therefore be considered a preliminary process for important detection and detection tasks.
- **Multi-task Learning:** The combination of multi-level backbone architecture features can be an essential step for improving detection performance. In addition, it can enhance performance to a great extent using more rich information by doing numerous computer vision tasks, such as object detection, semantic, and instance segmentation. Adopting this technique provides an efficient way for researchers to combine several functions in a model, thus enhancing detection precision without compromising processing speed.
- **GAN Object Detectors:** As we know, a deep learning object detector often requires enormous quantities of data for training purposes. In contrast, GAN object detectors are an influential structure that produces fake images. The combination of real-world scenarios and GAN simulated data helps detectors more robust and more generalized.

#### 5.2.2. *Future directions for Deep domain adaptive object detection (DDAOD)*

- **Combining the merits of adaptation methods:** A promising solution is that different adaptive category methods, such as [25], combined with stylistic transfer and robust pseudo-labeling, achieve more remarkable performance, should be added to the promise. A possible combination is to train a detector adversely and create pseudo labels for target samples with a trained detector.
- **Local nature detection:** Exploring the local nature of detection is another promising direction. For example, generating instance-level simulation samples similar to target domain instance-level samples and then synthesizing samples for detection using image instance-level patches and target domain background images.
- **Homogenous DDAOD:** Homogenic DDAOD is one of the most revised works, while heterogeneous DDAOD is more challenging because of a wider field gap. More research, such as adapting from the visible domain to the thermal infrared domain with large quantities of data, and is annotated data collection, is considered expensive worthy work. But work in this direction is expected to have a high impact.
- Finally, it is also promising to use a state-of-the-art adaptive domain classification model and integrate it into detection frameworks and explore domain shift detection from scratch.

#### 5.2.3. *Future directions for tiny object detection*

- **Transmission of information:** The context around small object instances plays a vital role in detecting small objects. However, contextual information for small object detection is not always useful. GBDNet [154] uses LSTM to control the transmission of different regional information, preventing unserviceable background noises. Further work on research will be how to effectively handle the transmission of contextual information?
- **Small object detection techniques weakly monitored:** Current small object detectors with a more profound knowledge utilize fully supervised models from well-noted pictures with bounding boxes and segmentation masks. However, the notation process of fully managed learning takes a long time and is ineffective. Weakly supervised learning involves using a few fully annotated images to detect many unannotated images, but complete supervision without fully labeled training data is not scalable. It is easy and highly effective for small object detectors to be trained only with annotations of object class but not with bounding images. The development of minor object detection algorithms based on poorly monitored learning is essential for further research.
- **Benchmarks and datasets evolving for small object detection:** Although popular datasets like the MS-COCO contain several smaller classes of objects, a large part of an image occupies several instances of the objects in the "small" category. We do not know enough about how difficult the detection task for small objects is or the existing sensor’s function. We need large datasets specifically for small-scale object detection to evaluate the performance of small-scale object detection algorithms, just as ImageNet [13] image classification dataset is used to recognizes action. Thus, it is a research direction for small objects that sets small objects and corresponding benchmarks.
- **Joint multi-task learning and improvement:** As everyone knows, it can be easier to detect small objects by combining several small object detection methods [section 3]. In addition, the simultaneous adoption of multiple visualization tasks (such as object detection, semantic segmentation, instance segmentation, etc.) can increase individual tasks' performance with a high degree of margin due to rich data. RDSNet Wang et al. designs the two-stream structure for learning both object-level characteristics (i.e., bounding boxes) and pixel-level characteristics (i.e., masks instance). Good results have been achieved on the MS-COCO dataset by the RDSNet, which combines object detection and instance segmentation. That means that multiple tasks are learned and optimized, which is an excellent way to connect various functions in a network. For future research, it is also a priority to make effective use of multi-task joint learning and optimization to improve the performance of small objects.
- **Small object detection framework:** It is now a paradigm for emerging model weights on large image classification datasets for object detection problems. However, it may not be an optimal solution due to existing conflicts between detection and classification tasks.

Most detection algorithms are based on or altered from the backbone classification networks, and only a few of them try various selections (such as CornerNet [338] and ExtremeNet [342] based on the Hourglass network). So how to develop a new framework that detects small objects directly is also a significant field for future research.

## ACKNOWLEDGMENTS

I am thankful to Prof Xi Li wholeheartedly, who supported me throughout this research journey and helped me with complete devotion and motivation. Being my supervisor, he performed an excellent role in analyzing different researches and guiding all issues throughout the research journey.